\documentclass{article} % For LaTeX2e
\usepackage{iclr2025_conference,times}

\usepackage{amsmath,amsfonts,bm}

\def\eqref#1{equation~\ref{#1}}
\def\1{\bm{1}}

\DeclareMathAlphabet{\mathsfit}{\encodingdefault}{\sfdefault}{m}{sl}
\SetMathAlphabet{\mathsfit}{bold}{\encodingdefault}{\sfdefault}{bx}{n}

\usepackage{epsfig}
\usepackage{graphicx}
\usepackage{amsmath}
\usepackage{amssymb}
\usepackage{caption}
\usepackage{multirow}
\usepackage{url}
\usepackage[table]{xcolor}
\usepackage{booktabs}
\usepackage{tabularx}

\usepackage[pagebackref=true,breaklinks=true,colorlinks,bookmarks=false]{hyperref}

\definecolor{Gray}{gray}{0.94}
\newlength\savewidth

\title{studentSplat :Your Student Model Learns Single-view 3D Gaussian Splatting}

\author{Yimu Pan\thanks{ Work done during an internship at Amazon.} \\
The Pennsylvania State University \\
\texttt{ymp5078@psu.edu} \\
\And
Hongda Mao, Qingshuang Chen, Yelin Kim\\
Amazon\\
\texttt{\{hongdam,cicichen,kimyelin\}@amazon.com} \\
}

\iclrfinalcopy % Uncomment for camera-ready version, but NOT for submission.
\begin{document}
\maketitle
\begin{figure}[ht!]
    \centering
    \includegraphics[width=0.9\textwidth]{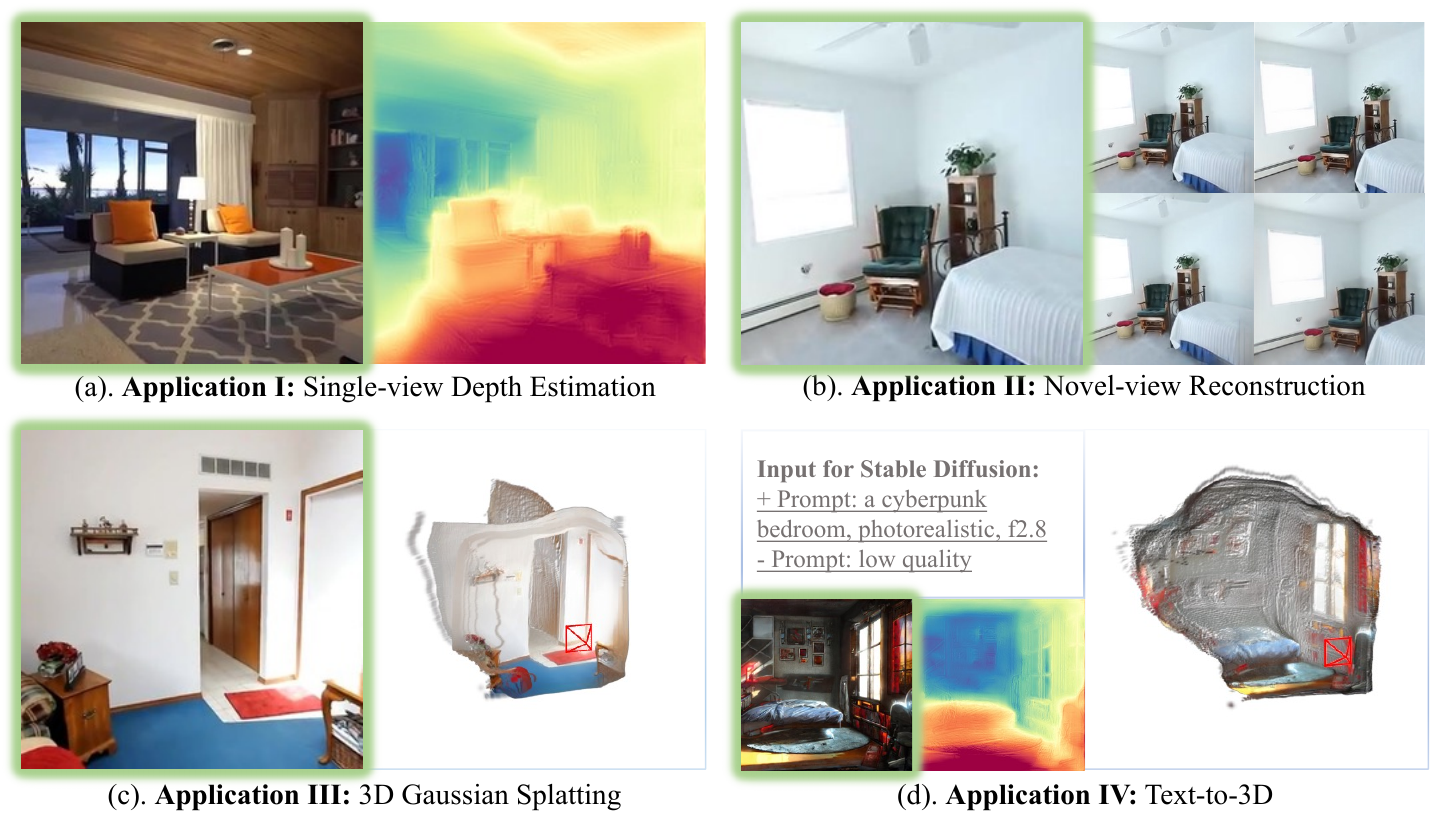}
    \caption{What can you do with \textbf{studentSplat}? All the results are generated using our studentSplat with teacher refine (detailed in the Appendix~\ref{sec:teacher_refine}) with only one input image. The input to our studentSplat is highlighted in green. studentSplat directly takes the generated image from Stable Diffusion~\citep{rombach2022high} in text-to-3D application.}
    \label{fig:demo}
\end{figure}

\begin{abstract}
Recent advance in feed-forward 3D Gaussian splatting has enable remarkable multi-view 3D scene reconstruction or single-view 3D object reconstruction but single-view 3D scene reconstruction remain under-explored due to inherited ambiguity in single-view. We present \textbf{studentSplat}, a single-view 3D Gaussian splatting method for scene reconstruction. To overcome the scale ambiguity and extrapolation problems inherent in novel-view supervision from a single input, we introduce two techniques: 1) a teacher-student architecture where a multi-view teacher model provides geometric supervision to the single-view student during training, addressing scale ambiguity and encourage geometric validity; and 2) an extrapolation network that completes missing scene context, enabling high-quality extrapolation. Extensive experiments show studentSplat achieves state-of-the-art single-view novel-view reconstruction quality and comparable performance to multi-view methods at the scene level. Furthermore, studentSplat demonstrates competitive performance as a self-supervised single-view depth estimation method, highlighting its potential for general single-view 3D understanding tasks.
\end{abstract}

\section{Introduction} 3D reconstruction is an essential task in robotics~\citep{yandun2020visual,han2022scene}, navigation~\citep{davison2003real,kazerouni2022survey}, virtual reality~\citep{bruno20103d}, and content creation~\citep{jun2023shap,tang2023dreamgaussian}. Advances in deep learning have enabled remarkable progress in 3D reconstruction~\citep{sitzmann2019scene,mildenhall2021nerf,truong2023sparf,kerbl20233d} through per-scene optimization using a large number of views. Recently, efficient feed-forward methods~\citep{yu2021pixelnerf,charatan2024pixelsplat} have been proposed to take a sparse set of input views and construct the 3D reconstruction, greatly improving efficiency. However, these methods require not only multi-view input but also the corresponding camera poses. Obtaining accurate camera poses usually involves a time and computation-intensive pipeline~\citep{ullman1979interpretation} and a large number of camera views or additional specialized networks~\citep{kendall2015posenet,yin2018geonet,peng2019pvnet}, which hinders the efficiency of feed-forward sparse-view 3D reconstruction. Single-view 3D reconstruction relaxes the requirements on both multi-view input and camera poses, serving as a more generalized alternative. Due to the inherent ambiguity in single-view input, current single-view 3D reconstruction works~\citep{yu2021pixelnerf,szymanowicz2024splatter} only operate at the object level.

In this work, we aim to expand single-view 3D object reconstruction to the scene level and propose a model capable of performing single-view 3D scene reconstruction using only multi-view supervision (i.e., no ground truth 3D annotations). In addition to the generalizability improvements from this extension, single-view 3D scene reconstruction holds the potential to perform self-supervised single-view vision tasks such as single-view depth estimation~\citep{li2018megadepth} and aid semantic segmentation~\citep{zhang2010semantic,schon2023impact}. Finally, a single-view 3D scene reconstruction model can be applied to the results from a text-to-image generation model~\citep{rombach2022high} to achieve text-to-3D scene generation without separate training.

To enable single-view 3D reconstruction, we adopt the 3D Gaussian splatter (3DGS)~\citep{kerbl20233d} representation. We identify and address two main problems in single-view 3DGS: scale ambiguity and extrapolation. We tackle these problems by proposing \textbf{studentSplat}, a single-view 3DGS model at the scene level. Since the unknown scale can be inferred when at least two input views are provided but is impossible using one input view~\citep{charatan2024pixelsplat,chen2024mvsplat}, our core design is to use a multi-view teacher model to estimate the 3D structure up to a scale and supervise the single-view student model using the teacher's estimation. Moreover, unlike a multi-view model that can bound the novel views by the input view frustums, a single-view model is required to extrapolate due to occlusion and camera view changes, which can lead to distortion of the 3DGS. We propose an extrapolator to complete the missing context in renderings before computing the novel-view reconstruction loss, both performing extrapolation and minimizing distortion. Extensive experiments show that our method can achieve good 3DGS on different benchmarks. Additionally, our method has the potential to connect 3DGS to self-supervised single-view vision tasks by demonstrating comparable performance to a self-supervised single-view depth estimation method.

Our contributions are summarized as follows: \begin{itemize} 
\item Propose the a single-view 3D scene Gaussian splatting model that does not require relative camera poses during inference. 
\item Address the extrapolation problem in single-view 3D scene reconstruction, which reduces distortion and produces out-of-context regions. 
\item Bridge the gap between multi-view 3D Gaussian splatting and self-supervised single-view depth estimation, expanding the applications of 3D Gaussian splatting models. 
% \item Enable text-to-3D scene generation, which is challenging for current methods. 
\end{itemize}

\section{Related Work}
\subsection{3D Representation}
Numerous 3D representations have been proposed to accommodate different applications. Point clouds are used in many applications~\citep{ullman1979interpretation,schoenberger2016sfm,nichol2022point} where the geometric shape is important. Recently, Neural Radiance Field (NeRF)~\citep{mildenhall2021nerf} is proposed to learn a view-based rendering function from multi-view supervision, but this learned function does not directly represent the geometric shape. 3DGS~\citep{kerbl20233d} is an efficient alternative representation similar to point clouds. Additionally, the efficient differentiable rendering implementation of 3DGS enables direct optimization of point clouds (3D Gaussians). This representation allows us to connect novel-view reconstruction to geometric reconstruction in an end-to-end manner.

\subsection{Feed-forward Multi-view 3D Reconstruction}
NeRF~\citep{mildenhall2021nerf} is one of the most popular representations for multi-view 3D reconstruction. PixelNeRF~\citep{yu2021pixelnerf} and GRF~\citep{trevithick2021grf} were among the earlier works that used a feed-forward network to produce radiance fields. Subsequent approaches improved rendering performance by incorporating cross-view feature matching~\citep{chen2021mvsnerf,chen2023explicit,du2023learning}, geometric encoding~\citep{miyato2023gta}, or target view information~\citep{xu2024murf}. Different from the predefined NeRF function, SRT~\citep{sajjadi2022scene} used a transformer to represent the rendering function. Another line of work closely related to ours was initiated by pixelSplat~\citep{charatan2024pixelsplat}, which directly predicted 3D Gaussians from multi-view images. latentSplat~\citep{wewer2024latentsplat} improved rendering performance by operating in the latent space, while MVSplat~\citep{chen2024mvsplat} incorporated cost-volume to improve both efficiency and performance. In contrast to previous approaches, our method requires only one input view, greatly improving the generalizability and versatility of the 3DGS model. Additionally, our method connects multi-view 3DGS to single-view vision tasks by learning one model that works on both tasks.

\subsection{Feed-forward Single-view 3D Reconstruction}
Single-view 3D reconstruction usually works at the object level. Unlike their multi-view counterparts, single-view 3D reconstruction requires extrapolation. Therefore, generative methods like diffusion models~\citep{rombach2022high,liu2023zero,tang2023dreamgaussian,liu2023syncdreamer} are used to complete the reconstruction. Similar to multi-view 3D reconstruction, radiance fields are popular among the rendering methods~\citep{liu2023zero,qian2023magic123,xu2023neurallift,tang2023make,melas2023realfusion,liu2024one}. TARS~\citep{duggal2022topologically} learns the deformation between 2D images and 3D objects. Recently, more approaches~\citep{szymanowicz2024splatter,tang2023dreamgaussian} have started using 3DGS for 3D reconstruction. Many other approaches~\citep{nichol2022point,jun2023shap} directly supervise the network using 3D object annotations. Additionally, some methods use learns directly from ground truth 3D annotation~\citep{yin2021learning,piccinelli2024unidepth}.
All existing work in single-view 3D reconstruction either requires 3D supervision or only works at the object level. In contrast, our method not only works at the scene level without 3D annotations but also has the potential to aid single-view vision tasks. 

\section{Our Approach}
\begin{figure}[ht!]
    \centering
    \includegraphics[width=0.85\textwidth]{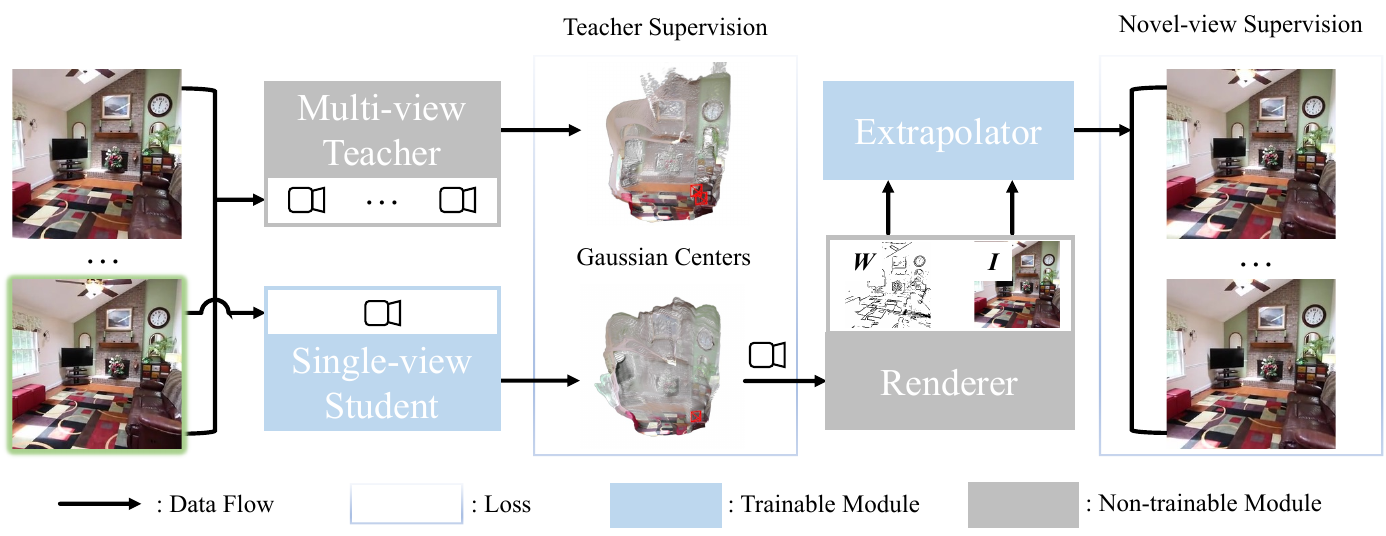}
    \caption{\textbf{The training pipeline of studentSplat.} The multi-view teacher network is used during training to produce 3D Gaussians centers (up-to an unknown scale) for geometric supervision. The input to student model is highlighted in green. The rendered student output is processed through the Extrapolator before performing novel-view supervision.}
    \label{fig:main}
\end{figure}
The overall pipeline is shown in Figure~\ref{fig:main}. We employ a multi-view 3DGS teacher network to provide geometric supervision, novel views to provide photometric supervision, and an extrapolator network to complete the missing context.

% \subsection{Preliminary}
% This section is about the problem (scale ambiguity) in single-view 3D reconstruction and how the problem has not been addressed in single-view but has solution in multi-view: 
% given the one camera view, we need at least one additional types of information from depth or scale to reconstruct the scene. Depth can be obtained using epipolar features if another view is provided. 

\subsection{Feed-forward 3D Gaussian Splatting}

In the multi-view 3DGS, we have $K$ sparse-view images $\mathcal{I}=\{{\bm I}^{i}\}_{i=1}^K$, (${\bm I}^i \in \mathbb{R}^{H \times W \times 3}$) and their corresponding camera projection matrices $\mathcal{P}=\{{\bm P}^i \}_{i=1}^K$, ${\bm P}^i=\mathbf{K}^i[\mathbf{R}^i|\mathbf{t}^i]$ where $\mathbf{K}^i$, $\mathbf{R}^i$ and $\mathbf{t}^i$ are the intrinsic, rotation, and translation matrices.
A multi-view 3DGS model $f^{K}_{\bm \theta}$, where $K$ is the number of views, maps images to 3D Gaussian parameters using
\begin{equation}
    f^{K}_{\bm \theta}: \{ ({\bm I}^{i}, {\bm P}^i) \}_{i=1}^K \mapsto \{(\bm{\mu}^j, \alpha^j, \bm{\Sigma}^j, \bm{c}^j )\}^{H \times W \times K}_{j=1}.
\end{equation}
On the other hand, the relaxed version, the single-view 3DGS model $f^1_{\bm \theta}$, performs the following: 
\begin{equation}
    f^1_{\bm \theta}: {\bm I^i} \mapsto \{(\bm{\mu}^j, \alpha^j, \bm{\Sigma}^j, \bm{c}^j )\}^{H \times W \times 1}_{j=1}.
\end{equation}

Unlike the multi-view 3DGS model, the single-view 3DGS model $f^1_{\bm \theta}$ is more prone to scale ambiguity and extrapolation issues. To train our studentSplat, we use both geometric and photometric supervisions: 
\begin{equation}
\mathcal{L}_{studentSplat}=\underbrace{\mathcal{L}_{geo}+\mathcal{L}_{grad}}_{\text{Teacher supervision}}+\underbrace{\mathcal{L}_{photo}}_{\text{Novel-view supervision}}.
\end{equation}
The following sections will explain how we address these issues and design each loss function.

\subsection{Teacher-student Model}

The aim of the teacher-student model is to solve the scale ambiguity problem during training time to enable single-view 3DGS for the student model with valid 3D geometric structure. Unlike their multi-view counterparts, a single-view model only accepts one view, making it difficult for the model itself to estimate the correct relative scale without cross-view feature matching and triangulation. 
% Taking inspiration from single-view depth estimation, we can train the student model to perform depth estimation and use the depth information to estimate the Gaussian centers.

\textbf{Using the teacher model geometric supervision.} Unlike previous approaches~\citep{nichol2022point,piccinelli2024unidepth}, we do not have access to ground truth 3D annotations. Despite the lack of 3D annotations, during training time, multiple views are provided, and cross-view feature matching can be performed to estimate the Gaussian center for each pixel in the context view with an implicit relative scale~\citep{charatan2024pixelsplat}. Thus, using the teacher model $f^{K}_{\bm \theta}$, we can convert the dateset from $\{ ({\bm I}^{i}, {\bm P}^i) \}_{i=1}^N$ to $\{ ({\bm I}^{i}, {\bm P}^i, {\bm{\mu}_t}^i) \}_{i=1}^N$. Then, in addition to the photometric loss computed from the target view $\{ ({\bm I}^{j}, {\bm P}^j) \}_{j=1}^K$, we supervise the student model's Gaussian center predictions $\bm{\mu}_s^i$ using the teacher's Gaussian centers $ \bm{\mu}_t^i$: $\mathcal{L}_{geo}=\lambda_{geo}\|\bm{\mu}_t^i-\bm{\mu}_s^i\|$, where $\|\cdot\|$ is the L1 loss.

\textbf{Regularizing local structure consistency.} The L1 loss used in $\mathcal{L}{geo}$ lacks consideration of the local structure which is prone to distortions in the less confident region such as the boundaries between the in- and out-of context region. To construct good 3D Gaussians and minimize distortions, we also need to encourage consistency in the local structure. Following previous work~\citep{li2018megadepth} that matches the depth map gradients to the ground truth depth map, we match the gradients of the teacher and student Gaussian centers using $\mathcal{L}_{grad}=\lambda_{grad}\|\nabla_{3D}\bm{\mu}_t^i-\nabla_{3D}\bm{\mu}_s^i\|$. Unlike previous work \citep{li2018megadepth} that defines the depth difference between nearby pixels as the gradient map (i.e., only the $z$ value is used for gradient computation), we propose a new definition of gradient $\nabla_{3D}$ that uses the 3D Euclidean distance (i.e., all $x$, $y$, and $z$ values are used for gradient computation) between nearby pixels as the gradient to accommodate 3D structure. This new definition is better aligned with 3D gradient matching.

\textbf{Discussion.}
The teacher model estimates only the relative scale, and consequently, the student model operates on the same relative scale. The teacher model will not be used at inference time. Therefore, we only require one input view, in other words, we relax the requirement for multiple input views and their corresponding camera poses, to perform the 3D reconstruction, which greatly improves the generalizability. More importantly, the resulting model naturally works as a single-view depth estimation model, connecting the 3D reconstruction task to single-view vision tasks, which goes beyond the capabilities of the teacher model.

\subsection{Extrapolation} 
Unlike multi-view scenarios where the photometric novel view reconstruction loss can be formulated using interpolation only (i.e., enclosing the novel camera view inside the context camera view frustums), single-view 3D reconstruction inevitably needs to extrapolate when computing the novel view reconstruction loss. This extrapolation can lead to distortion in the extrapolating region as there is no direct visual information. In the case of 3DGS, some 3D Gaussians will be forced to cover the extrapolation region to minimize the photometric loss, which compromises the geometric validity.

\textbf{Extrapolating the missing context.}
Although the teacher supervision will encourage the Gaussian centers to represent valid geometric shapes, the missing region will create a large photometric loss, which encourages spurious relationships. To minimize this unnecessary photometric loss, we need to either mask out the missing context during loss computation or fill the extrapolating region with additional pixels to avoid noisy gradient flow. We select the latter approach to achieve two functionalities: 1) guide the photometric loss to the correct Gaussians to minimize spurious relationships, and 2) perform extrapolation on the missing context to improve the novel-view reconstruction. We repurpose techniques from~\citep{luo2018single,ruckert2022adop} to achieve these functionalities. Instead of directly supervising the rasterized novel view $\hat{\bm{I}}^j=\mathrm{Rastrizer}(\mathbf{P}^j|\bm{\mu}^i, \alpha^i, \bm{\Sigma}^i, \bm{c}^i)$ , we further process the novel-view reconstructions through a network $g^{1}_{\bm \theta}$ and supervise the output $g^{1}_{\bm \theta}(\hat{\bm{I}}^j)$ using the photometric loss $\mathcal{L}_{photo}=\lambda_{l2}\|g^{1}_{\bm \theta}(\hat{\bm{I}}^j)-\bm I^j\|_2+\lambda_{lpips}\mathrm{LPIPS}(g^{1}_{\bm \theta}(\hat{\bm{I}}^j),\bm I^j)$, where $\|\cdot\|_2$ is the L2 loss and $\mathrm{LPIPS}$ is the Learned Perceptual Image Patch Similarity~\citep{zhang2018unreasonable} computed using VGG~\citep{simonyan2014very} features.

\textbf{Using composition to guide gradient flows.}
Directly applying $g^{1}_{\bm \theta}$ will prevent the rasterizer from getting direct supervision, which can harm the reconstruction quality. The ideal situation is to separate the missing context and the visible context using a confidence weight matrix $\bm W$ and treat their losses differently, but this separation is unknown before obtaining the 3D reconstruction. However, we can estimate the missing context using alpha compositing of the 3DGS. More specifically, we construct $\bm W$ by composing the $\alpha^i$. Intuitively, the missing context is less visible and has lower $\alpha^i$ whereas the visible context should have $\alpha^i=1$. We compose the novel view as $\hat{\bm{I}}^j_c=g^{1}_{\bm \theta}(\hat{\bm{I}}^j) \odot (\bm 1 - \bm W^j) + \hat{\bm{I}}^j \odot \bm W^j$, where $\bm W^j=\mathrm{Rastrizer}(\mathbf{P}^j|\bm{\mu}^i, \alpha^i, \bm{\Sigma}^i, \bm{1})$. Then, we can guide the gradients computed from $\mathcal{L}_{photo}=\lambda_{l2}\|\hat{\bm{I}}^j_c-\bm I^j\|_2+\lambda_{lpips}\mathrm{LPIPS}(\hat{\bm{I}}^j_c,\bm I^j)$ for the missing context to the extrapolation network, but the gradients for the context to the rasterizer, and the rasterizer always gets direct supervision from the reconstruction loss.  Additionally, the existence of $\bm W$ allows the student model to balance between the completeness and the confidence of the reconstruction by generating lower opacity for the regions with less confidence, since $g^{1}{\bm \theta}$ can still fill the less opaque area to minimize the loss. On the other hand, $\bm W$ cannot collapse to zero, as $g^{1}{\bm \theta}$ will not be able to fill anything without context. Finally, the learned $\bm W$ can be used during inference to identify the missing context.

\textbf{Discussion.}
Better extrapolation networks, such as diffusion-based methods~\citep{lkwq007_stablediffusion-infinity}, can be applied to achieve better novel-view reconstruction quality, but they make the training pipeline more complicated. We choose a feed-forward network (i.e., a pre-trained GAN network) to match the base training pipeline and preserve efficiency. The main goal of the extrapolator here is to direct the gradient flow to minimize artifacts. The ability to learn $\bm W$ is more important than generating the best extrapolation result; as long as some level of extrapolation can be achieved and the learned context mask $\bm W$ is accurate, we can apply more elaborate extrapolation methods, such as differential diffusion~\citep{levin2023differential}, during inference using the generated context mask. We visualize $\bm W$ in the Appendix~\ref{sec:add_results}. Because of the introduction of the extrapolator, we can use the student network to produce additional views by providing fake camera poses. This way, assuming the teacher model performs better than the student model, we can use the teacher model to process the student model's output views to further improve the reconstruction result. This setting is detailed in the Appendix~\ref{sec:teacher_refine}.
% ~\footnote{https://huggingface.co/blog/OzzyGT/outpainting-differential-diffusion}

\section{Experiments}
\subsection{Settings}
\textbf{Datasets.} To evaluate the novel-view reconstruction performance, we follow previous multi-view approaches~\citep{charatan2024pixelsplat,chen2024mvsplat} by using RealEstate10k (RE10k)~\citep{zhou2018stereo} and ACID~\citep{liu2021infinite}. These two datasets contain multiple views and the corresponding camera poses generated using a Structure from motion algorithm~\citep{schoenberger2016sfm} for different indoor and outdoor scenes. To evaluate the geometric quality and the potential to serve as a self-supervised depth estimation method, we use the indoor and outdoor annotations from DA-2K~\citep{yang2024depth} and DIODE~\citep{vasiljevic2019diode}.

\textbf{Metrics.} The novel-view reconstruction performance is evaluated using photometric metrics, including pixel-level Peak Signal-To-Noise Ratio (PSNR), patch-level Structural Similarity Index Measure (SSIM)~\citep{wang2004image}, and feature-level Learned Perceptual Image Patch Similarity 
 (LPIPS)~\citep{zhang2018unreasonable}. The depth estimation metrics follow standard practice by using Absolute Relative Error (AbsRel), $\delta_1$, and accuracy on the corresponding datasets. All experiments are performed using $256\times 256\times K$, where $K$ is the number of views. Thus, single-view methods have lower resolution. The evaluation settings are detailed in the Appendix~\ref{sec:add_eval_setting}.

\textbf{Implementation Details.} Since our goal is to design a new proof-of-concept approach instead of improving current ones, we aim for a balance between performance and efficiency rather than absolute performance. We expect larger models to produce better results. We use an efficient method, MVSplat~\citep{chen2024mvsplat}, as the teacher model. For the student model, we combine the DINOv2~\citep{oquab2023dinov2} pre-trained ViT-S backbone with the DPT~\citep{ranftl2021vision} head as the architecture, as it has been shown to perform well in single-view depth estimation. Following MVSplat, we use a shallow ResNet~\citep{he2016deep} encoded features and the original images to refine the output depth map. For the extrapolator, we use the efficient MI-GAN~\citep{sargsyan2023mi} inpainter. Additional details and results from different encoders are provided in the Appendix~\ref{sec:add_results}.

\subsection{Quantitative Comparisons}

\begin{table*}[t!]
    \begin{center}
\footnotesize
    \setlength{\tabcolsep}{2.pt} %
    \begin{tabular}{l@{}l c c ccc c ccc@{}}
    \toprule
    &\multirow{2}{*}[0pt]{Method} & \multirow{2}{*}[0pt]{\begin{tabular}[x]{@{}c@{}}Views\\(\#) \end{tabular}} & \multirow{2}{*}[0pt]{\begin{tabular}[x]{@{}c@{}}Params\\(M) \end{tabular}} & \multicolumn{3}{c}{RE10K~\scriptsize{\cite{zhou2018stereo}}} && \multicolumn{3}{c}{ACID~\scriptsize{\cite{liu2021infinite}}}\\
    \addlinespace[-12pt] \\
    \cmidrule{5-7} \cmidrule{9-11} 
    \addlinespace[-12pt] \\
    && & & PSNR$\uparrow$ & SSIM$\uparrow$ & LPIPS$\downarrow$ && PSNR$\uparrow$ & SSIM$\uparrow$ & LPIPS$\downarrow$ \\
    \midrule
    \parbox[t]{4mm}{\multirow{6}{*}{\rotatebox[origin=c]{90}{\textbf{\textcolor{lightgray}{Interpolation}}}}}
    & \textcolor{lightgray}{pixelNeRF}~\scriptsize{\cite{yu2021pixelnerf}} & \textcolor{lightgray}{2} & \textcolor{lightgray}{28.2} & \textcolor{lightgray}{20.43} & \textcolor{lightgray}{0.589} & \textcolor{lightgray}{0.550} && \textcolor{lightgray}{20.97} & \textcolor{lightgray}{0.547} & \textcolor{lightgray}{0.533} \\
     & \textcolor{lightgray}{GPNR}~\scriptsize{\cite{suhail2022generalizable}} & \textcolor{lightgray}{2} & \textcolor{lightgray}{9.6} & \textcolor{lightgray}{24.11} & \textcolor{lightgray}{0.793} & \textcolor{lightgray}{0.255} && \textcolor{lightgray}{25.28} & \textcolor{lightgray}{0.764} & \textcolor{lightgray}{0.332} \\
     & \textcolor{lightgray}{AttnRend}~\scriptsize{\cite{du2023learning}} & \textcolor{lightgray}{2} & \textcolor{lightgray}{125.1}  & \textcolor{lightgray}{24.78} & \textcolor{lightgray}{0.820} & \textcolor{lightgray}{0.213} && \textcolor{lightgray}{26.88} & \textcolor{lightgray}{0.799} & \textcolor{lightgray}{0.218} \\
     & \textcolor{lightgray}{MuRF}~\scriptsize{\cite{xu2024murf}} & \textcolor{lightgray}{2} & \textcolor{lightgray}{5.3}  & \textcolor{lightgray}{26.10} & \textcolor{lightgray}{0.858} & \textcolor{lightgray}{0.143} && \textcolor{lightgray}{28.09} & \textcolor{lightgray}{0.841} & \textcolor{lightgray}{0.155} \\
     & \textcolor{lightgray}{pixelSplat}~\scriptsize{\cite{charatan2024pixelsplat}} & \textcolor{lightgray}{2} & \textcolor{lightgray}{125.4}  & \textcolor{lightgray}{25.89} & \textcolor{lightgray}{0.858} & \textcolor{lightgray}{0.142} && \textcolor{lightgray}{28.14} & \textcolor{lightgray}{0.839} & \textcolor{lightgray}{0.150} \\
     & \textcolor{lightgray}{MVSplat}~\scriptsize{\cite{chen2024mvsplat}} &\textcolor{lightgray}{2} & \textcolor{lightgray}{12.0} & \textcolor{lightgray}{26.39} & \textcolor{lightgray}{0.869} & \textcolor{lightgray}{0.128} && \textcolor{lightgray}{28.25} & \textcolor{lightgray}{0.843} & \textcolor{lightgray}{0.144}  \\
    \midrule
    \parbox[t]{4mm}{\multirow{6}{*}{\rotatebox[origin=c]{90}{\textbf{Extrapolation}}}}
    % & pixelSplat~\cite{charatan2024pixelsplat} & 2 & 125.4  & - & - & - && - & - & - \\
     & pixelSplat~\scriptsize{\cite{charatan2024pixelsplat}} & 2 & 125.4  & {24.20} & {0.843} & {0.162} && {27.38} & 0.838 & {0.157} \\
     & {MVSplat}~\scriptsize{\cite{chen2024mvsplat}} & 2 & {12.0} & {23.48} & {0.834} & {0.163} && {26.39} & {0.831} & {0.158}  \\
     \cmidrule{2-11}
     & pixelSplat~\scriptsize{\cite{charatan2024pixelsplat}} & 1 & 125.4  & 20.15 & 0.662 & 0.256 && 23.40 & 0.670 & 0.242 \\
     & {MVSplat}~\scriptsize{\cite{chen2024mvsplat}} & 1 & \textbf{12.0} & {17.73} & {0.585} & {0.296} && {20.17} & {0.581} & {0.288}  \\
     & {SplatterImage}~\scriptsize{\cite{szymanowicz2024splatter}} & 1 & 62.1 & \underline{22.32} & \underline{0.754} & \underline{0.197} && \underline{25.08} & \underline{0.738} & \underline{0.204}  \\
    % \cmidrule{2-11}
     & \cellcolor{Gray}studentSplat & \cellcolor{Gray}1 & \cellcolor{Gray}\underline{32.0} & \cellcolor{Gray}\textbf{24.98} & \cellcolor{Gray}\textbf{0.794} & \cellcolor{Gray}\textbf{0.156} &\cellcolor{Gray}& \cellcolor{Gray}\textbf{26.94} & \cellcolor{Gray}\textbf{0.767} & \cellcolor{Gray}\textbf{0.160}  \\
     % & studentSplat-teacher (Ours) & 1 & 44.0 & {-} & {-} & {-} && {-} & {-} & {-}  \\
    \bottomrule
    \end{tabular}
    \end{center}
    \caption{\textbf{Novel-view reconstruction performance}. The best performance in the single-view setting is bold, the second is underlined. The original interpolation performance are included for reference.
    }
    \label{tab:sota_compare}
    \vspace{-0.4cm}
\end{table*}

\textbf{Novel-view reconstruction performance.} To perform a quantitative comparison with the current state-of-the-art (SOTA) methods on 3D scene reconstruction performance without 3D annotations, we follow previous work to evaluate the novel-view reconstruction. Additionally, we aim to evaluate the extrapolation capability. Therefore, unlike previous work that only bounded the novel views by the context view frustums, we also evaluate the reconstruction performance using views both inside and outside the context view frustums. As suggested by previous work~\citep{charatan2024pixelsplat} that current scene-level 3DGS methods cannot perform extrapolation, we can see from Table~\ref{tab:sota_compare} that the performance of a SOTA multi-view 3DGS method drops when performing extrapolation. Additionally, a single-view 3DGS method, SplatterImage~\citep{szymanowicz2024splatter}, outperforms the SOTA multi-view 3DGS method when only one view is provided, which suggests that multi-view 3DGS methods cannot be directly applied to the single-view setting despite their promising performance in the multi-view setting; directly training the single-view method will result in better reconstruction performance. This result supports the necessity of training a single-view 3DGS method. Furthermore, our studentSplat achieves the best single-view 3DGS performance and is on par with the multi-view models despite using only one input view, which demonstrates the effectiveness of the teacher-student architecture and extrapolation capability. However, we acknowledge that our SSIM score is still behind the multi-view methods. The inferior performance can be partially attributed to the resolution difference. All the methods generate one 3D Gaussian for each image pixel; the methods using two input views have twice the number of 3D Gaussians to render from, thus resulting in a sharper image, which in turn results in a higher SSIM score.

\begin{table*}[ht!]
    \begin{center}
\footnotesize
% \resizebox{\linewidth}{!}{
    \setlength{\tabcolsep}{2.pt} %
    \begin{tabular}{lcccccccccccccc}
    \toprule

    \multirow{2}{*}[0pt]{Method} &\multirow{2}{*}{\begin{tabular}[x]{@{}c@{}}Views\\(\#) \end{tabular}} & \multicolumn{3}{c}{ACID~\scriptsize{\citep{liu2021infinite}}} & \multicolumn{3}{c}{DTU~\scriptsize{\citep{aanaes2016large}}} \\
    % \addlinespace[-12pt] \\
    \cmidrule(lr){3-5} \cmidrule(lr){6-8} 
    \addlinespace[-12pt] \\
     & & PSNR$\uparrow$ & SSIM$\uparrow$ & LPIPS$\downarrow$ & PSNR$\uparrow$ & SSIM$\uparrow$ & LPIPS$\downarrow$ \\
    \midrule
    \textcolor{lightgray}{pixelSplat}~\scriptsize{\citep{charatan2024pixelsplat}}  & \textcolor{lightgray}{2} & \textcolor{lightgray}{27.64} & \textcolor{lightgray}{0.830} & \textcolor{lightgray}{0.160} & \textcolor{lightgray}{12.89} & \textcolor{lightgray}{0.382} & \textcolor{lightgray}{0.560} \\
    \textcolor{lightgray}{MVSplat}~\scriptsize{\citep{chen2024mvsplat}} & \textcolor{lightgray}{2} & \textcolor{lightgray}{28.15} & \textcolor{lightgray}{0.841} & \textcolor{lightgray}{0.147} & \textcolor{lightgray}{13.94} & \textcolor{lightgray}{0.473} & \textcolor{lightgray}{0.385}  \\
    \midrule
    {MVSplat}~\scriptsize{\citep{chen2024mvsplat}} & 1 & 21.13 & 0.631 & 0.261 & 9.67 & 0.245 & 0.602  \\
    {SplatterImage}~\scriptsize{\citep{szymanowicz2024splatter}} & 1 & \underline{24.95} & \underline{0.735} & \underline{0.200} & \underline{12.39} & \underline{0.353} & \underline{0.542}  \\
    \cellcolor{Gray}{studentSplat} & \cellcolor{Gray}1 & \cellcolor{Gray}\textbf{26.59} & \cellcolor{Gray}\textbf{0.758} & \cellcolor{Gray}\textbf{0.167} & \cellcolor{Gray}\textbf{14.15} & \cellcolor{Gray}\textbf{0.411} & \cellcolor{Gray}\textbf{0.491}  \\
                    
    \bottomrule
    \end{tabular}
    % }%
    \end{center}
    \caption{\textbf{Cross-dataset generalization in novel view reconstruction.}
 Results from models trained on RealEstate10K. The best performance in singel-view novel-view reconstruction is bold and the second is underlined. The original multi-view interpolation results are included for reference.}
    \label{tab:generalization}
    
    \vspace{-0.1in}
    
\end{table*}

\textbf{Novel-view reconstruction generalizability.} We follow MVSplat~\citep{chen2024mvsplat} to evaluate the cross-dataset novel-view reconstruction performance. As shown in Table~\ref{tab:generalization}, MVSplat again does not work in the single-view setting. On the other hand, our studentSplat achieved the best single-view performance and is on par with multi-view pixelSplat, depending on the dataset. This result further supports the effectiveness of our method and shows the potential for our method to act as a generalizable single-view vision encoder.

\begin{table}[h!]
    \begin{center}
\footnotesize
% \resizebox{0.6\linewidth}{!}{
    \setlength{\tabcolsep}{2.0pt} %
    \begin{tabular}{@{}lccc c@{}}
    \toprule
    \multirow{2}{*}[-2pt]{Method} & \multicolumn{2}{c}{DIODE~\scriptsize{\citep{vasiljevic2019diode}}}& & DA-2K~\scriptsize{\citep{yang2024depth}} \\
    \addlinespace[-12pt] \\
    \cmidrule{2-3} \cmidrule{5-5} 
    \addlinespace[-12pt] \\
    & $\delta_1$$\uparrow$ & AbsRel$\downarrow$ & & Acc (\%)$\uparrow$ \\
    \midrule
    % LeReS (Supervised)~\cite{yin2021learning}  & 0.751 & 0.287 && - \\
    % \midrule
    
    GasMono~\scriptsize{\citep{zhao2023gasmono}}  & \underline{0.504} & \textbf{0.348} && \underline{0.700} \\
    % pixelSplat~\cite{charatan2024pixelsplat}  & - & - && - \\
    % {MVSplat}~\cite{chen2024mvsplat}& - & - && -  \\
    {SplatterImage}~\scriptsize{\citep{szymanowicz2024splatter}} & 0.395 & 1.457 && 0.615  \\
    \cellcolor{Gray}{studentSplat} & \cellcolor{Gray}\textbf{0.604} & \cellcolor{Gray}\underline{0.407} &\cellcolor{Gray}& \cellcolor{Gray}\textbf{0.708} \\
    \bottomrule
    \end{tabular}
    % }%
    \end{center}
    \caption{\textbf{Cross-dataset generalization in self-supervised single-view depth estimation.} SplatterImage and studentSplat are trained on RealEstiate10K. GasMono is taken from the original work. Testing dataset unseen during training.
    }
    \label{tab:mono_depth}
\end{table}
\textbf{Self-supervised single-view depth estimation performance.} We evaluate the single-view depth estimation performance of our method against a SOTA self-supervised single-view depth estimation method, GasMono~\citep{zhao2023gasmono}, and a SOTA single-view object 3DGS model, SplatterImage~\citep{szymanowicz2024splatter}. Note that the evaluation datasets are unseen by any of the models. From Table~\ref{tab:mono_depth}, we see that our method achieved much better performance than SplatterImage and on-par performance with GasMono. This result further supports the generalizability of our method and the potential to serve as a self-supervised single-view depth estimation method.

\subsection{Qulitative Comparisons}
\begin{figure}[ht!]
    \centering
    \includegraphics[width=0.9\textwidth]{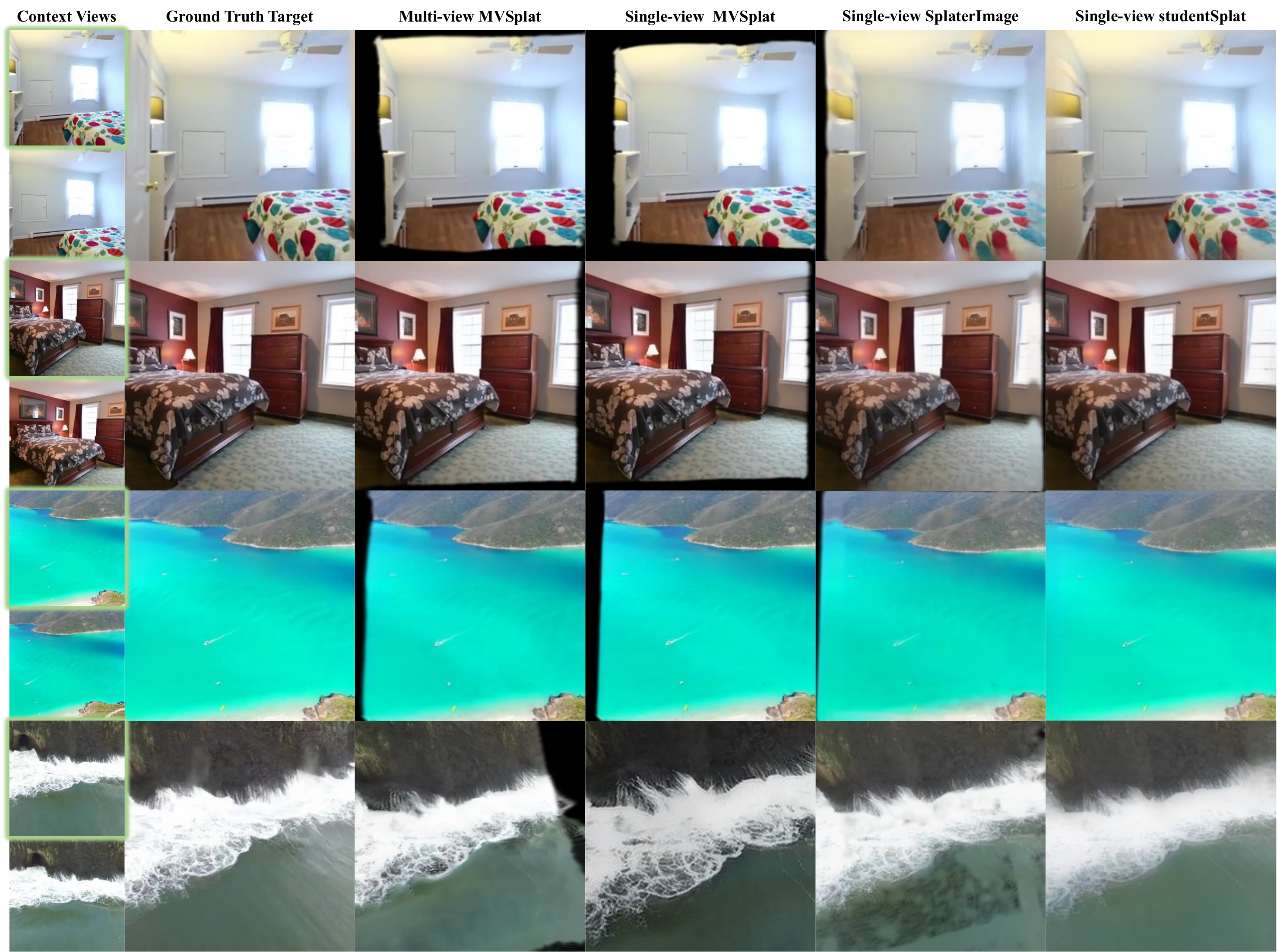}
    \caption{The qualitative comparison between representative methods in the extrapolation setting. Top two rows are from RE10K and the bottom two rows are from ACID The multi-view method use both of the context views whereas the single-view method only use the context view highlighted in green. Additional examples are in the Appendix~\ref{sec:add_results}.}
    \label{fig:re10k_acid_qualitative}
\end{figure}
In this section, we aim to visualize the proposed studentSplat in terms of extrapolation performance, distortion, and reconstruction quality. The qualitative comparisons for depth estimation and integration with Stable Diffusion~\citep{rombach2022high} for text-to-3D generation are in the Appendix~\ref{sec:add_results}.

\textbf{Better extrapolation performance.} Thanks to our extrapolator, our studentSplat is able to fill the missing context, as shown in Figure~\ref{fig:re10k_acid_qualitative}, whereas previous methods leave the region blank or stretch the border Gaussians to fill the region.

\textbf{Less distortion compared to current single-view methods.} From the last two columns of Figure~\ref{fig:re10k_acid_qualitative}, we see that SplatterImage tends to create a jelly effect around the border of the context, which is the distortion we aim to minimize, and our method does not have such distortion.

\textbf{Similar reconstruction quality with less resolution.} Since our studentSplat uses one input view instead of two views, we generate half the number of 3D Gaussians (i.e., half the resolution). Despite the lower resolution and sharpness, our studentSplat still generates overall comparable reconstructions to the multi-view (higher resolution) methods, as shown in Figure~\ref{fig:re10k_acid_qualitative}.

\begin{figure}[ht!]
    \centering
    \includegraphics[width=0.9\textwidth]{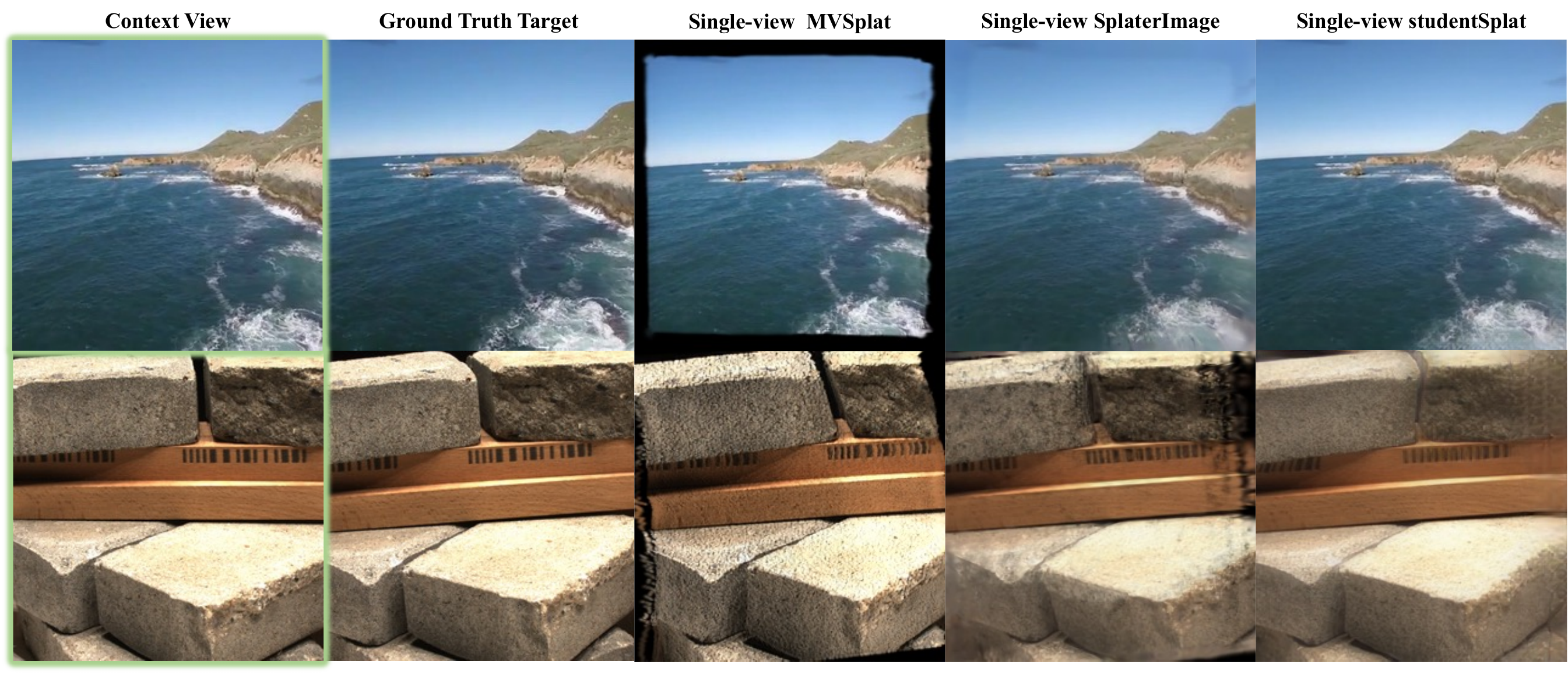}
    \caption{The qualitative comparison between representative methods in the single-view cross-dataset generalization setting. The context view is highlighted in green.}
    \label{fig:cross_qualitative}
\end{figure}

\textbf{Generalizable reconstruction quality.} The advantage of studentSplat generalizes to unseen domains. As shown in Figure~\ref{fig:cross_qualitative}, our method is able to complete the missing region with low distortion. However, due to the lower resolution (i.e., fewer 3D Gaussians), our results are less sharp.

\subsection{Ablation Study}
In this ablation study, we aim to evaluate how each of the proposed modules contributes to the model's performance by iteratively removing the proposed modules.

\textbf{Ablation on the extrapolator.} From Table~\ref{tab:ablation}, we observe that removing composition results in a slight performance drop across metrics. If we additionally remove the extrapolator, we have a large performance drop, which demonstrates the necessity of the extrapolator.

\begin{figure}[ht!]
    \centering
    \includegraphics[width=0.9\textwidth]{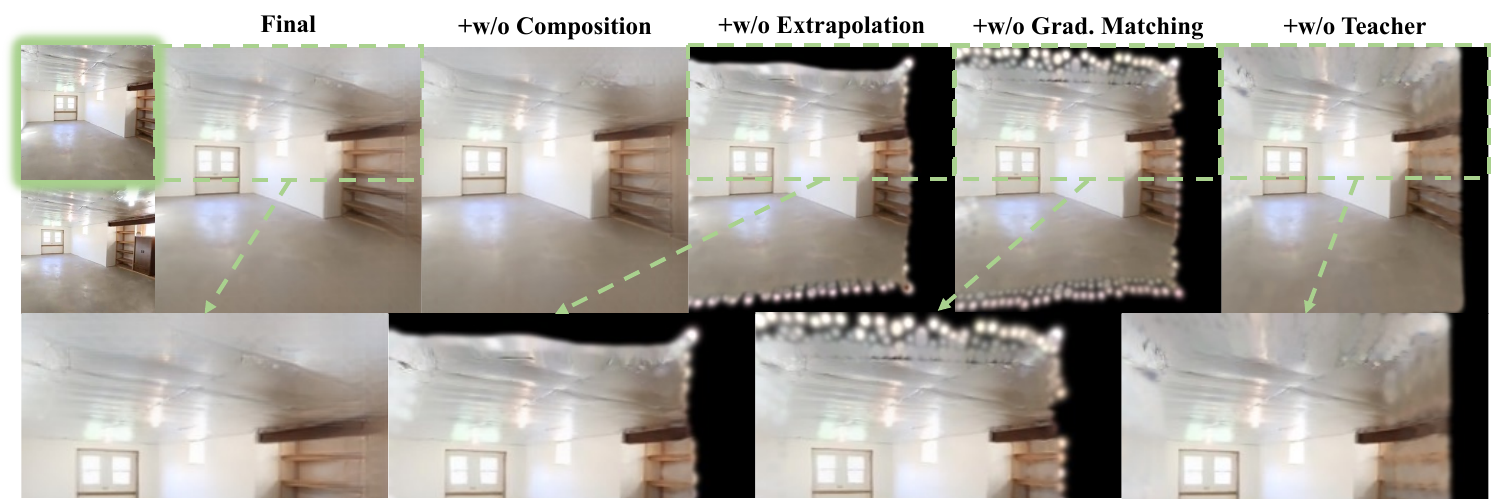}
    \caption{\textbf{The qualitative ablation results.} The input view is highlighted in green. The ground truth target view is below the input view. We zoomed in some areas for better comparison.}
    \label{fig:ablation}
\end{figure}

\textbf{Ablation on the teacher geometric supervision.} The quantitative measurements alone will be misleading for ablating the geometric supervision, as photometric measurements do not consider geometric validity. Therefore, we evaluate the modules both quantitatively and qualitatively. We see from Table~\ref{tab:ablation} that removing gradient matching improves PSNR but reduces LPIPS. This only makes sense when considering Figure~\ref{fig:ablation}, where removing gradient matching results in a large number of Gaussians being misplaced in the missing region. Although this misplacement improves the PSNR score, it lowers the structure validity which we aim to preserve. If we additionally remove the entire teacher geometric supervision, we observe that the reconstruction performance improves, which seems counterintuitive. However, we can see from Figure~\ref{fig:ablation} that the improvements again come with sacrificing the geometric validity; the jelly distortion similar to SplatterImage appears, which even affects the in-context region. Additional ablations in depth estimation performance are in the Appendix~\ref{sec:add_results}. These results demonstrate the effectiveness of the proposed modules.
\begin{table}[ht!]
\begin{center}
\footnotesize
% \resizebox{0.6\linewidth}{!}{
    \begin{tabular}{cllccccccccccc}
    \toprule
    Ablation Module &Setup  & PSNR$\uparrow$ & SSIM$\uparrow$ & LPIPS$\downarrow$ \\
    \midrule
     \cellcolor{Gray}&\cellcolor{Gray}Final & \cellcolor{Gray}24.98 & \cellcolor{Gray}0.794 & \cellcolor{Gray}0.156 \\
    \midrule
    \multirow{2}{*}{Extrapolator}&+w/o Composition & 24.85  & 0.792  & 0.158 \\
    &+w/o Extrapolation & 21.38 & 0.741 & 0.208 \\
    \midrule
    \multirow{2}{*}{Supervision}&+w/o Gradient Matching & 21.57  & 0.741  & 0.211 \\
    &+w/o Teacher & 22.13  & 0.757  & 0.195 \\
    \bottomrule
    \end{tabular}
\end{center}
    % }%
    \caption{\textbf{Ablations on RealEstate10K}. We separate the ablation into Extrapolator where we ablate the components in the extrapolator and Supervision where we ablate the geometric supervision loss. }
    \label{tab:ablation}
    
\end{table}

\subsection{Discussion and Conclusion}
We demonstrate, using studentSplat, the possibility of single-view 3DGS at scene level, bridging the gap between 3DGS and single-view depth estimation. With its modular design, studentSplat allows for versatile applications (see Appendix~\ref{sec:add_results} and ~\ref{sec:teacher_refine}) and easy incorporation of better modules.

\textbf{Limitations and future direction.} Our method relies on the teacher model, thus inheriting the limitations of the teacher model. It would be interesting to eliminate the need for the teacher model to further improve the capability of single-view scene-level 3DGS. Additionally, training a single-view 3DGS model is still more difficult than training its multi-view counterparts, so our method cannot outperform the multi-view method in its current stage. Large-scale training is an interesting direction to explore the capability of single-view 3DGS for both novel-view reconstruction and depth estimation tasks. Furthermore, we expect our method to also aid other vision tasks like semantic segmentation, which can be another direction to explore. Finally, as a proof-of-concept approach, many design aspects, such as model architecture and loss function design, can be optimized.



% \section{Reproducibility Statement}
% It is important that the work published in ICLR is reproducible. Authors are strongly encouraged to include a paragraph-long Reproducibility Statement at the end of the main text (before references) to discuss the efforts that have been made to ensure reproducibility. This paragraph should not itself describe details needed for reproducing the results, but rather reference the parts of the main paper, appendix, and supplemental materials that will help with reproducibility. For example, for novel models or algorithms, a link to a anonymous downloadable source code can be submitted as supplementary materials; for theoretical results, clear explanations of any assumptions and a complete proof of the claims can be included in the appendix; for any datasets used in the experiments, a complete description of the data processing steps can be provided in the supplementary materials. Each of the above are examples of things that can be referenced in the reproducibility statement. This optional reproducibility statement will not count toward the page limit, but should not be more than 1 page.

\bibliography{iclr2025_conference}

@inproceedings{chang2014learning,
  title={Learning spatial knowledge for text to 3D scene generation},
  author={Chang, Angel and Savva, Manolis and Manning, Christopher D},
  booktitle={Proceedings of the 2014 conference on empirical methods in natural language processing (EMNLP)},
  pages={2028--2038},
  year={2014}
}

@article{ouyang2023text2immersion,
  title={Text2immersion: Generative immersive scene with 3d gaussians},
  author={Ouyang, Hao and Heal, Kathryn and Lombardi, Stephen and Sun, Tiancheng},
  journal={arXiv preprint arXiv:2312.09242},
  year={2023}
}

@article{fridman2024scenescape,
  title={Scenescape: Text-driven consistent scene generation},
  author={Fridman, Rafail and Abecasis, Amit and Kasten, Yoni and Dekel, Tali},
  journal={Advances in Neural Information Processing Systems},
  volume={36},
  year={2024}
}

@inproceedings{zhang20243d,
  title={3D-SceneDreamer: Text-Driven 3D-Consistent Scene Generation},
  author={Zhang, Songchun and Zhang, Yibo and Zheng, Quan and Ma, Rui and Hua, Wei and Bao, Hujun and Xu, Weiwei and Zou, Changqing},
  booktitle={Proceedings of the IEEE/CVF Conference on Computer Vision and Pattern Recognition},
  pages={10170--10180},
  year={2024}
}

@article{zhang2024text2nerf,
  title={Text2nerf: Text-driven 3d scene generation with neural radiance fields},
  author={Zhang, Jingbo and Li, Xiaoyu and Wan, Ziyu and Wang, Can and Liao, Jing},
  journal={IEEE Transactions on Visualization and Computer Graphics},
  year={2024},
  publisher={IEEE}
}

@inproceedings{chen2021mvsnerf,
  title={Mvsnerf: Fast generalizable radiance field reconstruction from multi-view stereo},
  author={Chen, Anpei and Xu, Zexiang and Zhao, Fuqiang and Zhang, Xiaoshuai and Xiang, Fanbo and Yu, Jingyi and Su, Hao},
  booktitle={Proceedings of the IEEE/CVF international conference on computer vision},
  pages={14124--14133},
  year={2021}
}

@inproceedings{yu2021pixelnerf,
  title={pixelnerf: Neural radiance fields from one or few images},
  author={Yu, Alex and Ye, Vickie and Tancik, Matthew and Kanazawa, Angjoo},
  booktitle={Proceedings of the IEEE/CVF conference on computer vision and pattern recognition},
  pages={4578--4587},
  year={2021}
}

@article{ruckert2022adop,
  title={Adop: Approximate differentiable one-pixel point rendering},
  author={R{\"u}ckert, Darius and Franke, Linus and Stamminger, Marc},
  journal={ACM Transactions on Graphics (ToG)},
  volume={41},
  number={4},
  pages={1--14},
  year={2022},
  publisher={ACM New York, NY, USA}
}

@inproceedings{suhail2022generalizable,
  title={Generalizable patch-based neural rendering},
  author={Suhail, Mohammed and Esteves, Carlos and Sigal, Leonid and Makadia, Ameesh},
  booktitle={European Conference on Computer Vision},
  pages={156--174},
  year={2022},
  organization={Springer}
}

@inproceedings{sajjadi2022scene,
  title={Scene representation transformer: Geometry-free novel view synthesis through set-latent scene representations},
  author={Sajjadi, Mehdi SM and Meyer, Henning and Pot, Etienne and Bergmann, Urs and Greff, Klaus and Radwan, Noha and Vora, Suhani and Lu{\v{c}}i{\'c}, Mario and Duckworth, Daniel and Dosovitskiy, Alexey and others},
  booktitle={Proceedings of the IEEE/CVF Conference on Computer Vision and Pattern Recognition},
  pages={6229--6238},
  year={2022}
}

@article{chen2023explicit,
  title={Explicit correspondence matching for generalizable neural radiance fields},
  author={Chen, Yuedong and Xu, Haofei and Wu, Qianyi and Zheng, Chuanxia and Cham, Tat-Jen and Cai, Jianfei},
  journal={arXiv preprint arXiv:2304.12294},
  year={2023}
}

@inproceedings{du2023learning,
  title={Learning to render novel views from wide-baseline stereo pairs},
  author={Du, Yilun and Smith, Cameron and Tewari, Ayush and Sitzmann, Vincent},
  booktitle={Proceedings of the IEEE/CVF Conference on Computer Vision and Pattern Recognition},
  pages={4970--4980},
  year={2023}
}

@article{miyato2023gta,
  title={Gta: A geometry-aware attention mechanism for multi-view transformers},
  author={Miyato, Takeru and Jaeger, Bernhard and Welling, Max and Geiger, Andreas},
  journal={arXiv preprint arXiv:2310.10375},
  year={2023}
}

@inproceedings{xu2024murf,
  title={Murf: Multi-baseline radiance fields},
  author={Xu, Haofei and Chen, Anpei and Chen, Yuedong and Sakaridis, Christos and Zhang, Yulun and Pollefeys, Marc and Geiger, Andreas and Yu, Fisher},
  booktitle={Proceedings of the IEEE/CVF Conference on Computer Vision and Pattern Recognition},
  pages={20041--20050},
  year={2024}
}

@inproceedings{charatan2024pixelsplat,
  title={pixelsplat: 3d gaussian splats from image pairs for scalable generalizable 3d reconstruction},
  author={Charatan, David and Li, Sizhe Lester and Tagliasacchi, Andrea and Sitzmann, Vincent},
  booktitle={Proceedings of the IEEE/CVF Conference on Computer Vision and Pattern Recognition},
  pages={19457--19467},
  year={2024}
}

@inproceedings{szymanowicz2024splatter,
  title={Splatter image: Ultra-fast single-view 3d reconstruction},
  author={Szymanowicz, Stanislaw and Rupprecht, Chrisitian and Vedaldi, Andrea},
  booktitle={Proceedings of the IEEE/CVF Conference on Computer Vision and Pattern Recognition},
  pages={10208--10217},
  year={2024}
}

@article{wewer2024latentsplat,
  title={latentsplat: Autoencoding variational gaussians for fast generalizable 3d reconstruction},
  author={Wewer, Christopher and Raj, Kevin and Ilg, Eddy and Schiele, Bernt and Lenssen, Jan Eric},
  journal={arXiv preprint arXiv:2403.16292},
  year={2024}
}

@article{chen2024mvsplat,
  title={Mvsplat: Efficient 3d gaussian splatting from sparse multi-view images},
  author={Chen, Yuedong and Xu, Haofei and Zheng, Chuanxia and Zhuang, Bohan and Pollefeys, Marc and Geiger, Andreas and Cham, Tat-Jen and Cai, Jianfei},
  journal={arXiv preprint arXiv:2403.14627},
  year={2024}
}

@inproceedings{luo2018single,
  title={Single view stereo matching},
  author={Luo, Yue and Ren, Jimmy and Lin, Mude and Pang, Jiahao and Sun, Wenxiu and Li, Hongsheng and Lin, Liang},
  booktitle={Proceedings of the IEEE conference on computer vision and pattern recognition},
  pages={155--163},
  year={2018}
}

@inproceedings{trevithick2021grf,
  title={Grf: Learning a general radiance field for 3d representation and rendering},
  author={Trevithick, Alex and Yang, Bo},
  booktitle={Proceedings of the IEEE/CVF International Conference on Computer Vision},
  pages={15182--15192},
  year={2021}
}

@inproceedings{yin2021learning,
  title={Learning to recover 3d scene shape from a single image},
  author={Yin, Wei and Zhang, Jianming and Wang, Oliver and Niklaus, Simon and Mai, Long and Chen, Simon and Shen, Chunhua},
  booktitle={Proceedings of the IEEE/CVF Conference on Computer Vision and Pattern Recognition},
  pages={204--213},
  year={2021}
}

@inproceedings{duggal2022topologically,
  title={Topologically-aware deformation fields for single-view 3d reconstruction},
  author={Duggal, Shivam and Pathak, Deepak},
  booktitle={Proceedings of the IEEE/CVF Conference on Computer Vision and Pattern Recognition},
  pages={1536--1546},
  year={2022}
}

@inproceedings{xu2023neurallift,
  title={Neurallift-360: Lifting an in-the-wild 2d photo to a 3d object with 360deg views},
  author={Xu, Dejia and Jiang, Yifan and Wang, Peihao and Fan, Zhiwen and Wang, Yi and Wang, Zhangyang},
  booktitle={Proceedings of the IEEE/CVF Conference on Computer Vision and Pattern Recognition},
  pages={4479--4489},
  year={2023}
}

@inproceedings{tang2023make,
  title={Make-it-3d: High-fidelity 3d creation from a single image with diffusion prior},
  author={Tang, Junshu and Wang, Tengfei and Zhang, Bo and Zhang, Ting and Yi, Ran and Ma, Lizhuang and Chen, Dong},
  booktitle={Proceedings of the IEEE/CVF international conference on computer vision},
  pages={22819--22829},
  year={2023}
}

@inproceedings{melas2023realfusion,
  title={Realfusion: 360deg reconstruction of any object from a single image},
  author={Melas-Kyriazi, Luke and Laina, Iro and Rupprecht, Christian and Vedaldi, Andrea},
  booktitle={Proceedings of the IEEE/CVF conference on computer vision and pattern recognition},
  pages={8446--8455},
  year={2023}
}

@inproceedings{liu2023zero,
  title={Zero-1-to-3: Zero-shot one image to 3d object},
  author={Liu, Ruoshi and Wu, Rundi and Van Hoorick, Basile and Tokmakov, Pavel and Zakharov, Sergey and Vondrick, Carl},
  booktitle={Proceedings of the IEEE/CVF international conference on computer vision},
  pages={9298--9309},
  year={2023}
}

@article{tang2023dreamgaussian,
  title={Dreamgaussian: Generative gaussian splatting for efficient 3d content creation},
  author={Tang, Jiaxiang and Ren, Jiawei and Zhou, Hang and Liu, Ziwei and Zeng, Gang},
  journal={arXiv preprint arXiv:2309.16653},
  year={2023}
}

@article{qian2023magic123,
  title={Magic123: One image to high-quality 3d object generation using both 2d and 3d diffusion priors},
  author={Qian, Guocheng and Mai, Jinjie and Hamdi, Abdullah and Ren, Jian and Siarohin, Aliaksandr and Li, Bing and Lee, Hsin-Ying and Skorokhodov, Ivan and Wonka, Peter and Tulyakov, Sergey and others},
  journal={arXiv preprint arXiv:2306.17843},
  year={2023}
}

@article{liu2024one,
  title={One-2-3-45: Any single image to 3d mesh in 45 seconds without per-shape optimization},
  author={Liu, Minghua and Xu, Chao and Jin, Haian and Chen, Linghao and Varma T, Mukund and Xu, Zexiang and Su, Hao},
  journal={Advances in Neural Information Processing Systems},
  volume={36},
  year={2024}
}

@article{nichol2022point,
  title={Point-e: A system for generating 3d point clouds from complex prompts},
  author={Nichol, Alex and Jun, Heewoo and Dhariwal, Prafulla and Mishkin, Pamela and Chen, Mark},
  journal={arXiv preprint arXiv:2212.08751},
  year={2022}
}

@article{jun2023shap,
  title={Shap-e: Generating conditional 3d implicit functions},
  author={Jun, Heewoo and Nichol, Alex},
  journal={arXiv preprint arXiv:2305.02463},
  year={2023}
}

@inproceedings{piccinelli2024unidepth,
  title={UniDepth: Universal Monocular Metric Depth Estimation},
  author={Piccinelli, Luigi and Yang, Yung-Hsu and Sakaridis, Christos and Segu, Mattia and Li, Siyuan and Van Gool, Luc and Yu, Fisher},
  booktitle={Proceedings of the IEEE/CVF Conference on Computer Vision and Pattern Recognition},
  pages={10106--10116},
  year={2024}
}

@article{liu2023syncdreamer,
  title={Syncdreamer: Generating multiview-consistent images from a single-view image},
  author={Liu, Yuan and Lin, Cheng and Zeng, Zijiao and Long, Xiaoxiao and Liu, Lingjie and Komura, Taku and Wang, Wenping},
  journal={arXiv preprint arXiv:2309.03453},
  year={2023}
}

@article{russakovsky2015imagenet,
  title={Imagenet large scale visual recognition challenge},
  author={Russakovsky, Olga and Deng, Jia and Su, Hao and Krause, Jonathan and Satheesh, Sanjeev and Ma, Sean and Huang, Zhiheng and Karpathy, Andrej and Khosla, Aditya and Bernstein, Michael and others},
  journal={International journal of computer vision},
  volume={115},
  pages={211--252},
  year={2015},
  publisher={Springer}
}

@article{aanaes2016large,
  title={Large-scale data for multiple-view stereopsis},
  author={Aan{\ae}s, Henrik and Jensen, Rasmus Ramsb{\o}l and Vogiatzis, George and Tola, Engin and Dahl, Anders Bjorholm},
  journal={International Journal of Computer Vision},
  volume={120},
  pages={153--168},
  year={2016},
  publisher={Springer}
}

@article{zhou2018stereo,
  title={Stereo Magnification: Learning view synthesis using multiplane images},
  author={Zhou, Tinghui and Tucker, Richard and Flynn, John and Fyffe, Graham and Snavely, Noah},
  journal={ACM Trans. Graph},
  volume={37},
  year={2018}
}

@inproceedings{liu2021infinite,
  title={Infinite nature: Perpetual view generation of natural scenes from a single image},
  author={Liu, Andrew and Tucker, Richard and Jampani, Varun and Makadia, Ameesh and Snavely, Noah and Kanazawa, Angjoo},
  booktitle={Proceedings of the IEEE/CVF International Conference on Computer Vision},
  pages={14458--14467},
  year={2021}
}

@article{vasiljevic2019diode,
  title={Diode: A dense indoor and outdoor depth dataset},
  author={Vasiljevic, Igor and Kolkin, Nick and Zhang, Shanyi and Luo, Ruotian and Wang, Haochen and Dai, Falcon Z and Daniele, Andrea F and Mostajabi, Mohammadreza and Basart, Steven and Walter, Matthew R and others},
  journal={arXiv preprint arXiv:1908.00463},
  year={2019}
}

@inproceedings{zhang2018unreasonable,
  title={The unreasonable effectiveness of deep features as a perceptual metric},
  author={Zhang, Richard and Isola, Phillip and Efros, Alexei A and Shechtman, Eli and Wang, Oliver},
  booktitle={Proceedings of the IEEE conference on computer vision and pattern recognition},
  pages={586--595},
  year={2018}
}

@article{wang2004image,
  title={Image quality assessment: from error visibility to structural similarity},
  author={Wang, Zhou and Bovik, Alan C and Sheikh, Hamid R and Simoncelli, Eero P},
  journal={IEEE transactions on image processing},
  volume={13},
  number={4},
  pages={600--612},
  year={2004},
  publisher={IEEE}
}

@inproceedings{zhao2023gasmono,
  title={GasMono: Geometry-aided self-supervised monocular depth estimation for indoor scenes},
  author={Zhao, Chaoqiang and Poggi, Matteo and Tosi, Fabio and Zhou, Lei and Sun, Qiyu and Tang, Yang and Mattoccia, Stefano},
  booktitle={Proceedings of the IEEE/CVF International Conference on Computer Vision},
  pages={16209--16220},
  year={2023}
}

@inproceedings{ranftl2021vision,
  title={Vision transformers for dense prediction},
  author={Ranftl, Ren{\'e} and Bochkovskiy, Alexey and Koltun, Vladlen},
  booktitle={Proceedings of the IEEE/CVF international conference on computer vision},
  pages={12179--12188},
  year={2021}
}

@article{yang2024depth,
  title={Depth Anything V2},
  author={Yang, Lihe and Kang, Bingyi and Huang, Zilong and Zhao, Zhen and Xu, Xiaogang and Feng, Jiashi and Zhao, Hengshuang},
  journal={arXiv preprint arXiv:2406.09414},
  year={2024}
}

@article{simonyan2014very,
  title={Very deep convolutional networks for large-scale image recognition},
  author={Simonyan, Karen and Zisserman, Andrew},
  journal={arXiv preprint arXiv:1409.1556},
  year={2014}
}

@inproceedings{he2016deep,
  title={Deep residual learning for image recognition},
  author={He, Kaiming and Zhang, Xiangyu and Ren, Shaoqing and Sun, Jian},
  booktitle={Proceedings of the IEEE conference on computer vision and pattern recognition},
  pages={770--778},
  year={2016}
}

@inproceedings{sargsyan2023mi,
  title={Mi-gan: A simple baseline for image inpainting on mobile devices},
  author={Sargsyan, Andranik and Navasardyan, Shant and Xu, Xingqian and Shi, Humphrey},
  booktitle={Proceedings of the IEEE/CVF International Conference on Computer Vision},
  pages={7335--7345},
  year={2023}
}

@article{oquab2023dinov2,
  title={Dinov2: Learning robust visual features without supervision},
  author={Oquab, Maxime and Darcet, Timoth{\'e}e and Moutakanni, Th{\'e}o and Vo, Huy and Szafraniec, Marc and Khalidov, Vasil and Fernandez, Pierre and Haziza, Daniel and Massa, Francisco and El-Nouby, Alaaeldin and others},
  journal={arXiv preprint arXiv:2304.07193},
  year={2023}
}

@inproceedings{caron2021emerging,
  title={Emerging properties in self-supervised vision transformers},
  author={Caron, Mathilde and Touvron, Hugo and Misra, Ishan and J{\'e}gou, Herv{\'e} and Mairal, Julien and Bojanowski, Piotr and Joulin, Armand},
  booktitle={Proceedings of the IEEE/CVF international conference on computer vision},
  pages={9650--9660},
  year={2021}
}

@article{ullman1979interpretation,
  title={The interpretation of structure from motion},
  author={Ullman, Shimon},
  journal={Proceedings of the Royal Society of London. Series B. Biological Sciences},
  volume={203},
  number={1153},
  pages={405--426},
  year={1979},
  publisher={The Royal Society London}
}

@inproceedings{schoenberger2016sfm,
    author={Sch\"{o}nberger, Johannes Lutz and Frahm, Jan-Michael},
    title={Structure-from-Motion Revisited},
    booktitle={Conference on Computer Vision and Pattern Recognition (CVPR)},
    year={2016},
}

@article{sitzmann2019scene,
  title={Scene representation networks: Continuous 3d-structure-aware neural scene representations},
  author={Sitzmann, Vincent and Zollh{\"o}fer, Michael and Wetzstein, Gordon},
  journal={Advances in Neural Information Processing Systems},
  volume={32},
  year={2019}
}

@article{mildenhall2021nerf,
  title={Nerf: Representing scenes as neural radiance fields for view synthesis},
  author={Mildenhall, Ben and Srinivasan, Pratul P and Tancik, Matthew and Barron, Jonathan T and Ramamoorthi, Ravi and Ng, Ren},
  journal={Communications of the ACM},
  volume={65},
  number={1},
  pages={99--106},
  year={2021},
  publisher={ACM New York, NY, USA}
}

@inproceedings{truong2023sparf,
  title={Sparf: Neural radiance fields from sparse and noisy poses},
  author={Truong, Prune and Rakotosaona, Marie-Julie and Manhardt, Fabian and Tombari, Federico},
  booktitle={Proceedings of the IEEE/CVF Conference on Computer Vision and Pattern Recognition},
  pages={4190--4200},
  year={2023}
}

@article{kerbl20233d,
  title={3D Gaussian Splatting for Real-Time Radiance Field Rendering.},
  author={Kerbl, Bernhard and Kopanas, Georgios and Leimk{\"u}hler, Thomas and Drettakis, George},
  journal={ACM Trans. Graph.},
  volume={42},
  number={4},
  pages={139--1},
  year={2023}
}

@inproceedings{yandun2020visual,
  title={Visual 3d reconstruction and dynamic simulation of fruit trees for robotic manipulation},
  author={Yandun, Francisco and Silwal, Abhisesh and Kantor, George},
  booktitle={Proceedings of the IEEE/CVF Conference on Computer Vision and Pattern Recognition Workshops},
  pages={54--55},
  year={2020}
}

@article{han2022scene,
  title={Scene reconstruction with functional objects for robot autonomy},
  author={Han, Muzhi and Zhang, Zeyu and Jiao, Ziyuan and Xie, Xu and Zhu, Yixin and Zhu, Song-Chun and Liu, Hangxin},
  journal={International Journal of Computer Vision},
  volume={130},
  number={12},
  pages={2940--2961},
  year={2022},
  publisher={Springer}
}

@inproceedings{davison2003real,
  title={Real-time simultaneous localisation and mapping with a single camera},
  author={Davison},
  booktitle={Proceedings Ninth IEEE International Conference on Computer Vision},
  pages={1403--1410},
  year={2003},
  organization={IEEE}
}

@article{kazerouni2022survey,
  title={A survey of state-of-the-art on visual SLAM},
  author={Kazerouni, Iman Abaspur and Fitzgerald, Luke and Dooly, Gerard and Toal, Daniel},
  journal={Expert Systems with Applications},
  volume={205},
  pages={117734},
  year={2022},
  publisher={Elsevier}
}

@article{bruno20103d,
  title={From 3D reconstruction to virtual reality: A complete methodology for digital archaeological exhibition},
  author={Bruno, Fabio and Bruno, Stefano and De Sensi, Giovanna and Luchi, Maria-Laura and Mancuso, Stefania and Muzzupappa, Maurizio},
  journal={Journal of Cultural Heritage},
  volume={11},
  number={1},
  pages={42--49},
  year={2010},
  publisher={Elsevier}
}

@inproceedings{zhang2010semantic,
  title={Semantic segmentation of urban scenes using dense depth maps},
  author={Zhang, Chenxi and Wang, Liang and Yang, Ruigang},
  booktitle={Computer Vision--ECCV 2010: 11th European Conference on Computer Vision, Heraklion, Crete, Greece, September 5-11, 2010, Proceedings, Part IV 11},
  pages={708--721},
  year={2010},
  organization={Springer}
}

@inproceedings{li2018megadepth,
  title={Megadepth: Learning single-view depth prediction from internet photos},
  author={Li, Zhengqi and Snavely, Noah},
  booktitle={Proceedings of the IEEE conference on computer vision and pattern recognition},
  pages={2041--2050},
  year={2018}
}

@inproceedings{schon2023impact,
  title={Impact of pseudo depth on open world object segmentation with minimal user guidance},
  author={Sch{\"o}n, Robin and Ludwig, Katja and Lienhart, Rainer},
  booktitle={Proceedings of the IEEE/CVF Conference on Computer Vision and Pattern Recognition},
  pages={4809--4819},
  year={2023}
}

@inproceedings{rombach2022high,
  title={High-resolution image synthesis with latent diffusion models},
  author={Rombach, Robin and Blattmann, Andreas and Lorenz, Dominik and Esser, Patrick and Ommer, Bj{\"o}rn},
  booktitle={Proceedings of the IEEE/CVF conference on computer vision and pattern recognition},
  pages={10684--10695},
  year={2022}
}

@misc{lkwq007_stablediffusion-infinity, author = {lkwq007}, title = {stablediffusion-infinity}, year = {2023}, publisher = {GitHub}, journal = {GitHub repository}, howpublished = {\url{https://github.com/lkwq007/stablediffusion-infinity}}, }

@misc{levin2023differential,
      title={Differential Diffusion: Giving Each Pixel Its Strength},
      author={Eran Levin and Ohad Fried},
      year={2023},
      eprint={2306.00950},
      archivePrefix={arXiv},
      primaryClass={cs.CV}
}

@article{kingma2014adam,
  title={Adam: A method for stochastic optimization},
  author={Kingma, Diederik P and Ba, Jimmy},
  journal={arXiv preprint arXiv:1412.6980},
  year={2014}
}

@article{hendrycks2016gaussian,
  title={Gaussian error linear units (gelus)},
  author={Hendrycks, Dan and Gimpel, Kevin},
  journal={arXiv preprint arXiv:1606.08415},
  year={2016}
}

@inproceedings{kendall2015posenet,
  title={Posenet: A convolutional network for real-time 6-dof camera relocalization},
  author={Kendall, Alex and Grimes, Matthew and Cipolla, Roberto},
  booktitle={Proceedings of the IEEE international conference on computer vision},
  pages={2938--2946},
  year={2015}
}

@inproceedings{peng2019pvnet,
  title={Pvnet: Pixel-wise voting network for 6dof pose estimation},
  author={Peng, Sida and Liu, Yuan and Huang, Qixing and Zhou, Xiaowei and Bao, Hujun},
  booktitle={Proceedings of the IEEE/CVF conference on computer vision and pattern recognition},
  pages={4561--4570},
  year={2019}
}

@inproceedings{yin2018geonet,
  title={Geonet: Unsupervised learning of dense depth, optical flow and camera pose},
  author={Yin, Zhichao and Shi, Jianping},
  booktitle={Proceedings of the IEEE conference on computer vision and pattern recognition},
  pages={1983--1992},
  year={2018}
}

@article{eigen2014depth,
  title={Depth map prediction from a single image using a multi-scale deep network},
  author={Eigen, David and Puhrsch, Christian and Fergus, Rob},
  journal={Advances in neural information processing systems},
  volume={27},
  year={2014}
}
\bibliographystyle{iclr2025_conference}

\clearpage
\appendix
\section{Additional Implementation Details}
\label{sec:add_implement}
\begin{figure}[ht!]
    \centering
    \includegraphics[width=\textwidth]{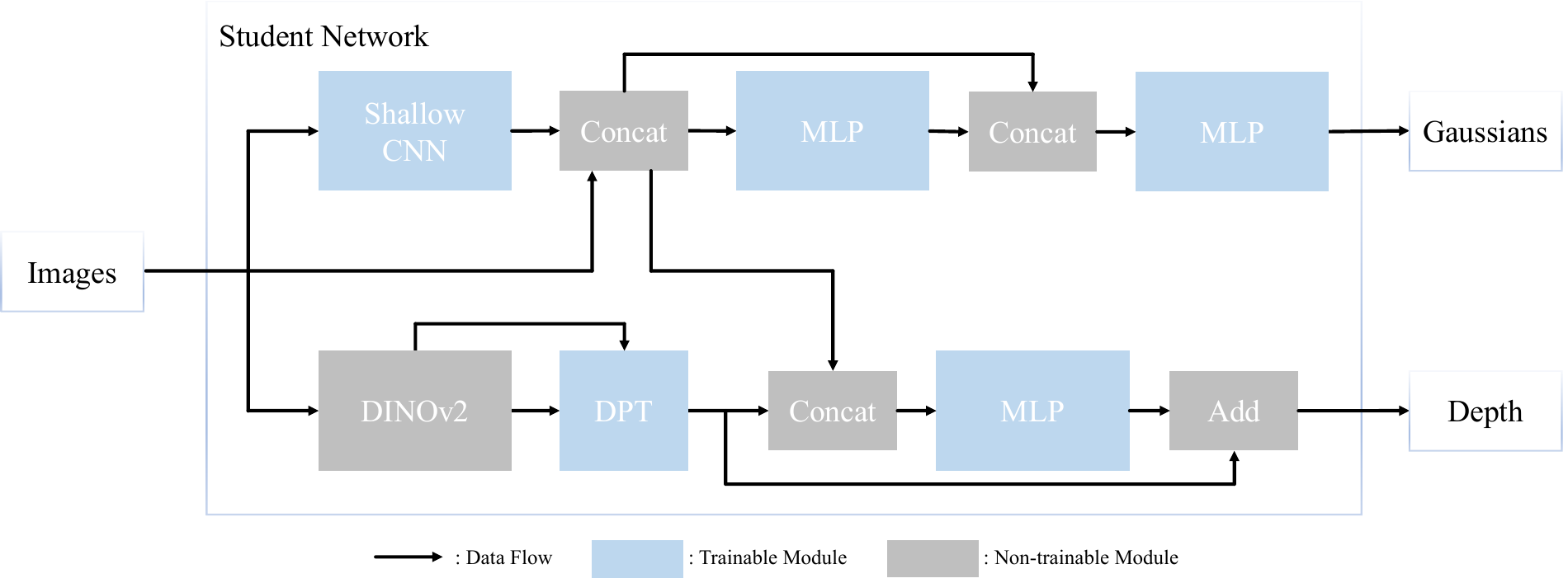}
    \caption{\textbf{Student network architecture.} The shallow CNN is the same as previous work~\citep{chen2024mvsplat} but randomly initialized. The MLP conposes of a 3x3 Conv, a GeLU~\citep{hendrycks2016gaussian} activation, and a 1x1 Conv. }
    \label{fig:student_arch}
\end{figure}

\textbf{Student Architecture.} The student network architecture is shown in Figure~\ref{fig:student_arch}. It only requires the images as input (i.e., without camera pose requirements). It comprises a backbone branch and a refine branch, similar to previous work~\citep{charatan2024pixelsplat,chen2024mvsplat}. The backbone branch localizes the Gaussian centers along the $z$-axis, whereas the refine branch uses CNN features and input images to refine the backbone prediction and predict other Gaussian parameters.

\textbf{Novel-view reconstruction.} To generate a novel-view, our studentSplat first generates the novel views directly using the rendering function from 3DGS. Additionally, we use the rendering function to generate the opacity map. The novel-view renderings and the opacity map are processed by the extrapolator to generate the complete novel views.

\textbf{Depth normalization.} To learn a generalizable depth map, we use the provided camera intrinsics, near plane, and far plane to scale, shift, and clip the predicted depth map, respectively, using: $\mathrm{depth\_scaled}=\mathrm{Max}(\mathrm{focal\_length}*\mathrm{depth}+\mathrm{near},\mathrm{far})$

\textbf{More training details.} All our models are trained on two A10G GPUs with a total batch size of 2 for 300,000 iterations with the Adam~\citep{kingma2014adam} optimizer. Each batch contains one training scene (i.e., two input views and four target views).  For all experiments, we use an initial learning rate of 2e-4 and a cosine learning rate scheduler with 2000 warm-up iterations. All the models are trained for 300,000 iterations. Same as MVSplat~\citep{chen2024mvsplat}, the frame distance between two input views is gradually increased as the training progresses. For both RE10K~\citep{zhou2018stereo} and ACID~\citep{liu2021infinite}, we follow previous works~\citep{charatan2024pixelsplat,chen2024mvsplat} to set the near and far depth planes to 1 and 100, respectively. For DTU~\citep{aanaes2016large}, we use the provided near and far depth planes of 2.125 and 4.525, respectively.

\section{Evaluation Settings}
\label{sec:add_eval_setting}

\textbf{Novel-view reconstruction.} For the interpolation setting, we use the reported numbers from previous work for reference. Those evaluations are done using 3 novel views inside the context frustums. In our extrapolation setting, we use 2 novel views outside and one novel view inside the context frustums. All the multi-view methods use all the context views to produce the 3D Gaussians. Comparing single-view methods to multi-view methods is inherently unfair since single-view methods have less information and lower resolutions (i.e., fewer 3D Gaussians). Although we cannot avoid this unfairness, to better compare multi-view and single-view methods, we use the context view that produces the best SSIM score for each target view as the input for the single-view method. It is not intuitive to apply the multi-view methods (i.e., pixelSplat~\citep{charatan2024pixelsplat} and MVSplat~\citep{chen2024mvsplat}) in the single-view setting. To adapt them, we simply repeat the input view to create another view, since the training data already contains views that are very close to each other. We noticed in the GitHub issue \url{https://github.com/donydchen/mvsplat/issues/37} that we may warp the input view to create a fake view. However, this is impossible without the scale or depth information.

\textbf{Depth estimation.} DA2K~\citep{yang2024depth} is annotated by human on depth relationship between two pixels (i.e., which pixel is closer). To make sure both pixels are on the same image and to keep the aspect ratio, we pad the shorter edge of the image to the longer edge size and resize to $256 \times 256$. DIODE\citep{vasiljevic2019diode} dataset has the ground truth depth map with mask, we first extract two square crops from each image with maximum coverage and resize each crop to $256 \times 256$. Next, we perform median scaling to both the predicted depth map and the ground truth depth map. Then, we apply the mask on both the predicted depth map and the ground truth depth map before computing the metrics. Finally, we average the metrics over all the crops.

\textbf{Depth estimation metric.}
The metrics are defined following previous work~\citep{eigen2014depth}. More specifically, the AbsRel, the absolute value of the difference between predicted depth and ground truth depth relative to the ground truth depth, and $\delta_1$, the percentage of pixel with predicted depth close enough to the ground truth depth, are defined as:

\begin{equation}
    \mathrm{AbsRel}(\hat{\bm D},\bm D)=\frac{1}{\|\bm D\|}\sum_{\hat{d},d\in \hat{\bm D},\bm D}|\hat{d}-d|/d,\; 
\end{equation}
\begin{equation}
    \delta_1(\hat{\bm D},\bm D) = \frac{1}{\|\bm D\|}\|\{\hat{d},d\in \hat{\bm D},\bm D |\mathrm{Max}(\frac{\hat{d}}{d},\frac{d}{\hat{d}}) < 1.25\}\|,
\end{equation}
where $|\cdot|$ is the absolute value, $\|\cdot\|$ is the size of a matrix of the cardinality of a set, $\bm D$ is the ground truth depth map, $\hat{\bm D}$ is the predicted depth map, and $\hat{d},d\in \hat{\bm D},\bm D$ represents taking the depth values $\hat{d},d$ from each matrix at the corresponding pixels.

\section{More Results}
\label{sec:add_results}
\textbf{Encoder without large-scale pre-training.} We also trained our model on RE10K dataset using DINO~\citep{caron2021emerging} with ImageNet~\citep{russakovsky2015imagenet} pre-trained weights (i.e., one tenth of the training data compare to DINOv2~\citep{oquab2023dinov2}) to evaluate how much the pre-trained encoder contributes to our model performance. From Table~\ref{tab:dinov1}, we observe a performance drop without using large-scale pre-trained weights which is expected. However, the performance drop is much smaller compare to model trained without the proposed modules. Therefore, the proposed modules are the main contributor to studentSplat's performance.
\begin{table}[ht!]
    \begin{center}
    \begin{tabular}{llccccccccccc}
    \toprule
    Setup  & PSNR$\uparrow$ & SSIM$\uparrow$ & LPIPS$\downarrow$ \\
    \midrule
    \cellcolor{Gray}Final & \cellcolor{Gray}24.98 & \cellcolor{Gray}0.794 & \cellcolor{Gray}0.156 \\
    +w/o Extrapolation & 21.38 & 0.741 & 0.208 \\
    +w/o Teacher & 22.13  & 0.757  & 0.195 \\
    \midrule
    w/o Large-scale Pre-train & 24.63  & 0.783  & 0.163 \\
    
    \bottomrule
    \end{tabular}
    \end{center}
    \caption{Compare novel view reconstruction results w/ and w/o large-scale pre-trained encoder on RealEstate10K}
    \label{tab:dinov1}
    \vspace{-0.1in}
    
\end{table}

\textbf{Student model with ground truth depth pre-training.} We also trained our model on the RE10K dataset using pre-trained weights from Depth Anything V2~\citep{yang2024depth} (i.e., the training data contains ground truth depth labels) to evaluate if we can enhance the reconstruction quality when the student model has prior depth knowledge. From Table~\ref{tab:depth_anything}, we observe a performance improvement using Depth Anything V2 weights, which suggests that the performance of our studentSplat can be further improved if we employ a depth estimation model as the student model. This result further reinforces the connection between depth estimation and 3DGS.
\begin{table}[ht!]
% \resizebox{\linewidth}{!}{

    \begin{center}
    \begin{tabular}{llccccccccccc}
    \toprule
    Setup  & PSNR$\uparrow$ & SSIM$\uparrow$ & LPIPS$\downarrow$ \\
    \midrule
    \cellcolor{Gray}Final & \cellcolor{Gray}24.98 & \cellcolor{Gray}0.794 & \cellcolor{Gray}0.156 \\
    +w/ Depth Anything V2 Weights & 25.11  & 0.798  & 0.154 \\
    
    \bottomrule
    \end{tabular}
    \end{center}
    % }%
    \caption{Compare novel view reconstruction results w/ and w/o Depth Anything V2~\protect\cite{yang2024depth} weights}
    \label{tab:depth_anything}
    % \vspace{-0.1in}
    
\end{table}

\textbf{Depth estimation and teacher supervision.} In addition to the ablation results in the main text, we validate the effectiveness of teacher supervision on geometric validity by performing depth estimation. As shown in Table~\ref{tab:mono_depth_teacher_sup_ablation}, the method without gradient matching performs worse, and the model without teacher supervision suffers a significant performance drop. These results further validate the effectiveness of the proposed teacher supervision.

\begin{table*}[ht!]
    \begin{center}
\footnotesize
    \setlength{\tabcolsep}{2.0pt} %
    \begin{tabular}{@{}lccc c@{}}
    \toprule
    \multirow{2}{*}[-2pt]{Method} & \multicolumn{2}{c}{DIODE~\scriptsize{\citep{vasiljevic2019diode}}}& & DA-2K~\scriptsize{\citep{yang2024depth}} \\
    \addlinespace[-12pt] \\
    \cmidrule{2-3} \cmidrule{5-5} 
    \addlinespace[-12pt] \\
    & $\delta_1$$\uparrow$ & AbsRel$\downarrow$ & & Acc (\%)$\uparrow$ \\
    \midrule
    % LeReS (Supervised)~\citep{yin2021learning}  & 0.751 & 0.287 && - \\
    % \midrule
    
    GasMono~\scriptsize{\citep{zhao2023gasmono}}  & {0.504} & {0.348} && {0.700} \\{SplatterImage}~\scriptsize{\citep{szymanowicz2024splatter}} & 0.395 & 1.457 && 0.615  \\
    \midrule
    \cellcolor{Gray}{Final} & \cellcolor{Gray}{0.604} & \cellcolor{Gray}{0.407} &\cellcolor{Gray}& \cellcolor{Gray}{0.708} \\
    {+w/o Gradient Matching} & {0.606} & {0.413} && {0.683} \\
    {+w/o Teacher} & {0.541} & {1.526} && {0.653} \\
    \bottomrule
    \end{tabular}
    \end{center}
    \caption{Cross-dataset generalization in self-supervised single-view depth estimation w/ and w/o teacher supervision.
    }
    \label{tab:mono_depth_teacher_sup_ablation}
    \vspace{-0.4cm}
\end{table*}

\begin{figure}[t!]
    \centering
    \includegraphics[width=\textwidth]{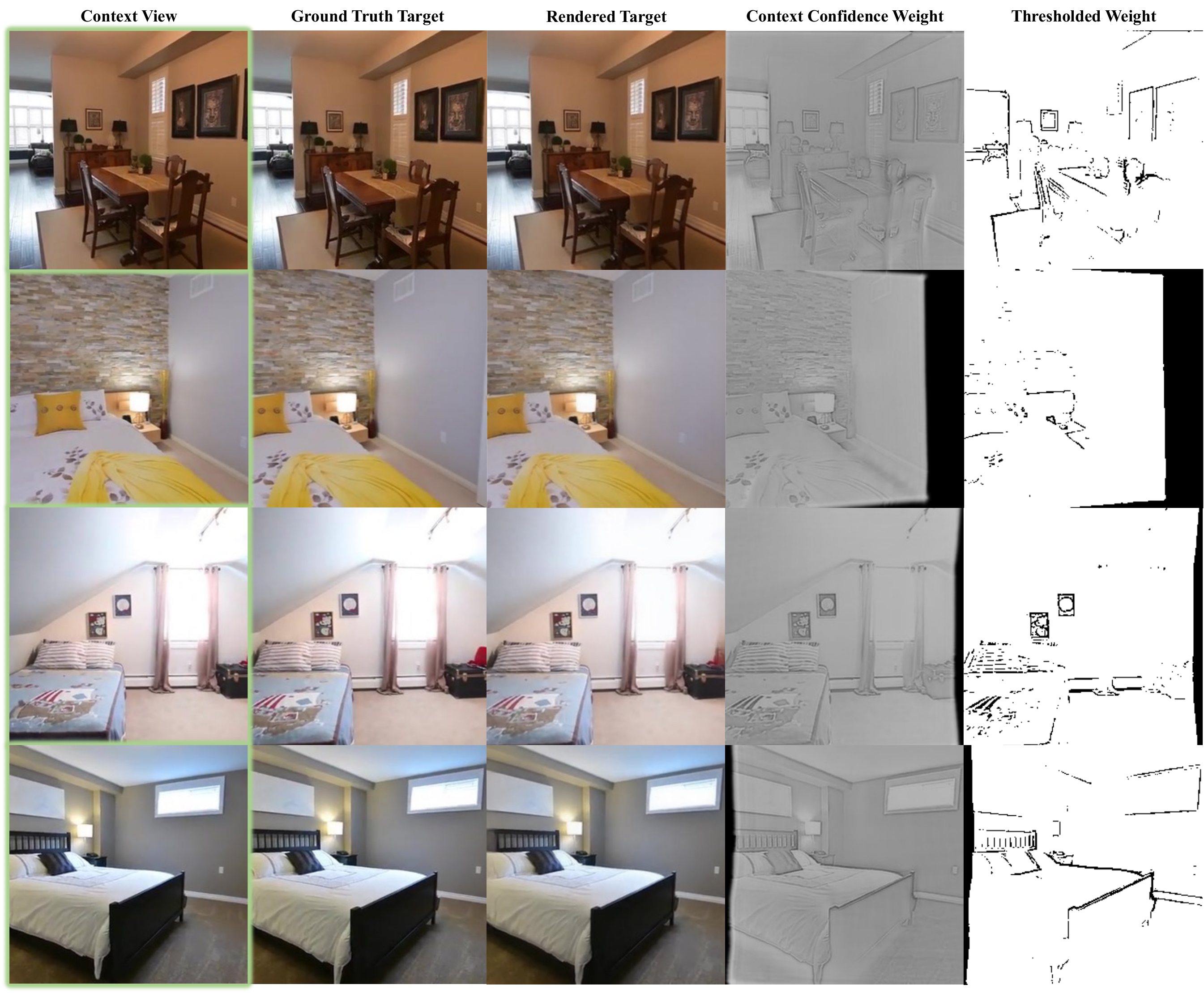}
    \caption{Visualization of the context confidence weight $\bm {W}$ on RE10K dataset. Our studentSplat is more confident at the brighter regions and less confident at the darker regions. The less confident regions of the rendered target are complete by the extrapolator.}
    \label{fig:confidence_weight_mat}
\end{figure}

\begin{figure}[t!]
    \centering
    \includegraphics[width=\textwidth]{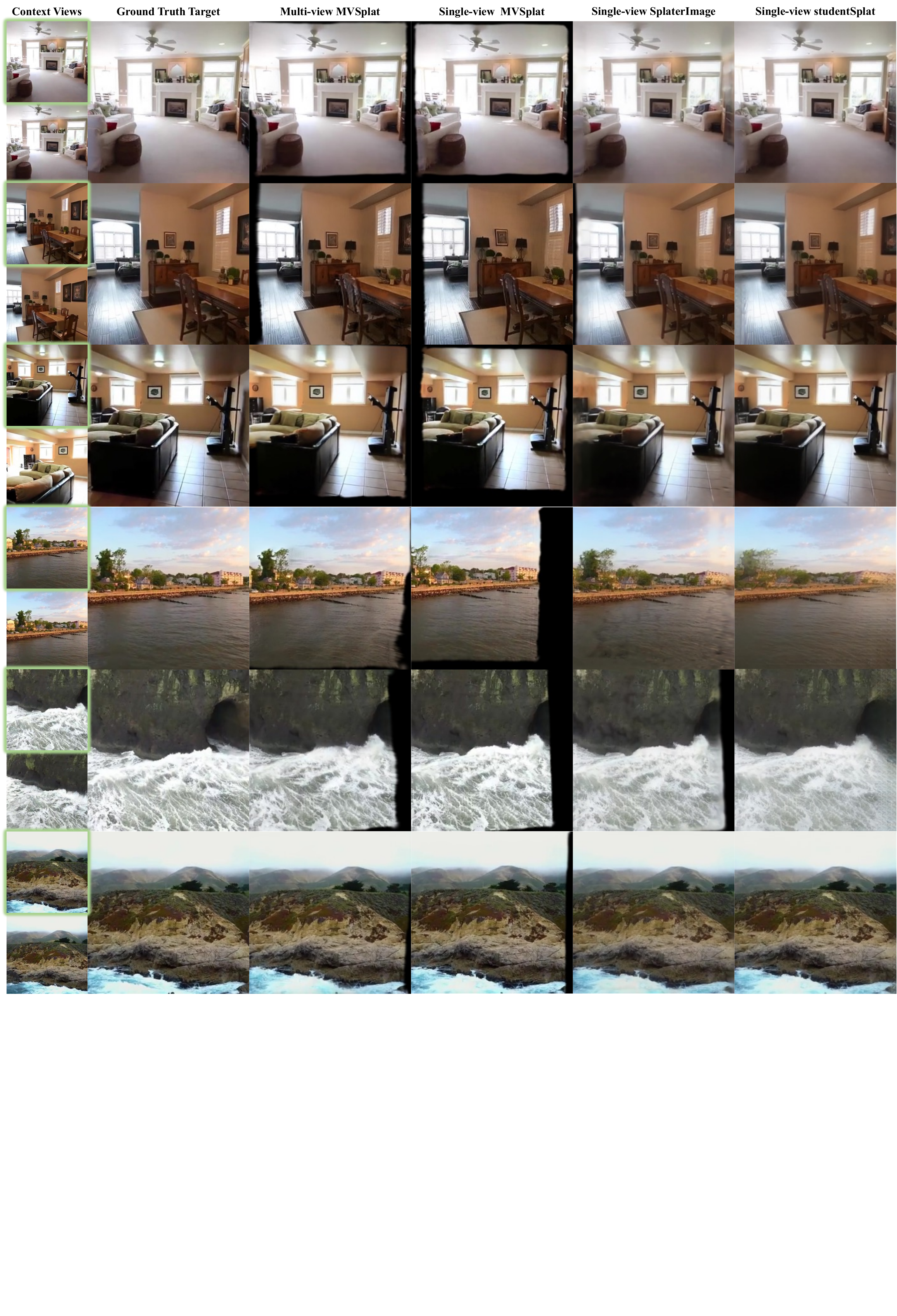}
    \caption{Additional qualitative comparison between representative methods in the extrapolation setting. The top four rows are from RE10K, and the bottom four rows are from ACID. The multi-view method uses both context views, whereas the single-view method only uses the context view highlighted in green.}
    \label{fig:add_re10k_acid_qualitative}
\end{figure}

\begin{figure}[t!]
    \centering
    \includegraphics[width=0.9\textwidth]{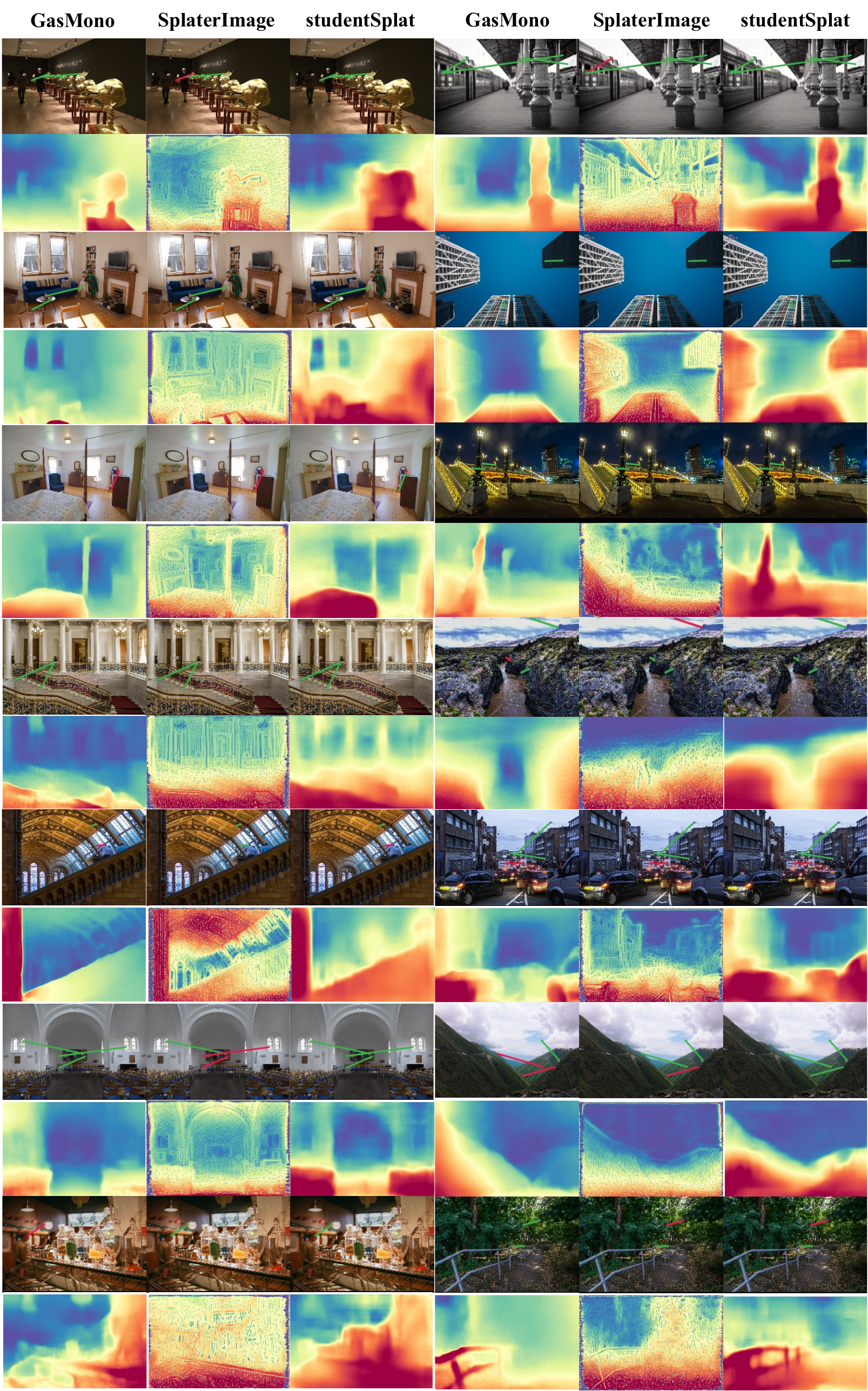}
    \caption{Additional qualitative comparison between representative methods for self-supervised single-view depth estimation performance on the DA2K dataset. Line segments in the original images represent the predicted depth difference (red: incorrect, green: correct).}
    \label{fig:da2k_qualitative}
\end{figure}

\begin{figure}[t!]
    \centering
    \includegraphics[width=0.9\textwidth]{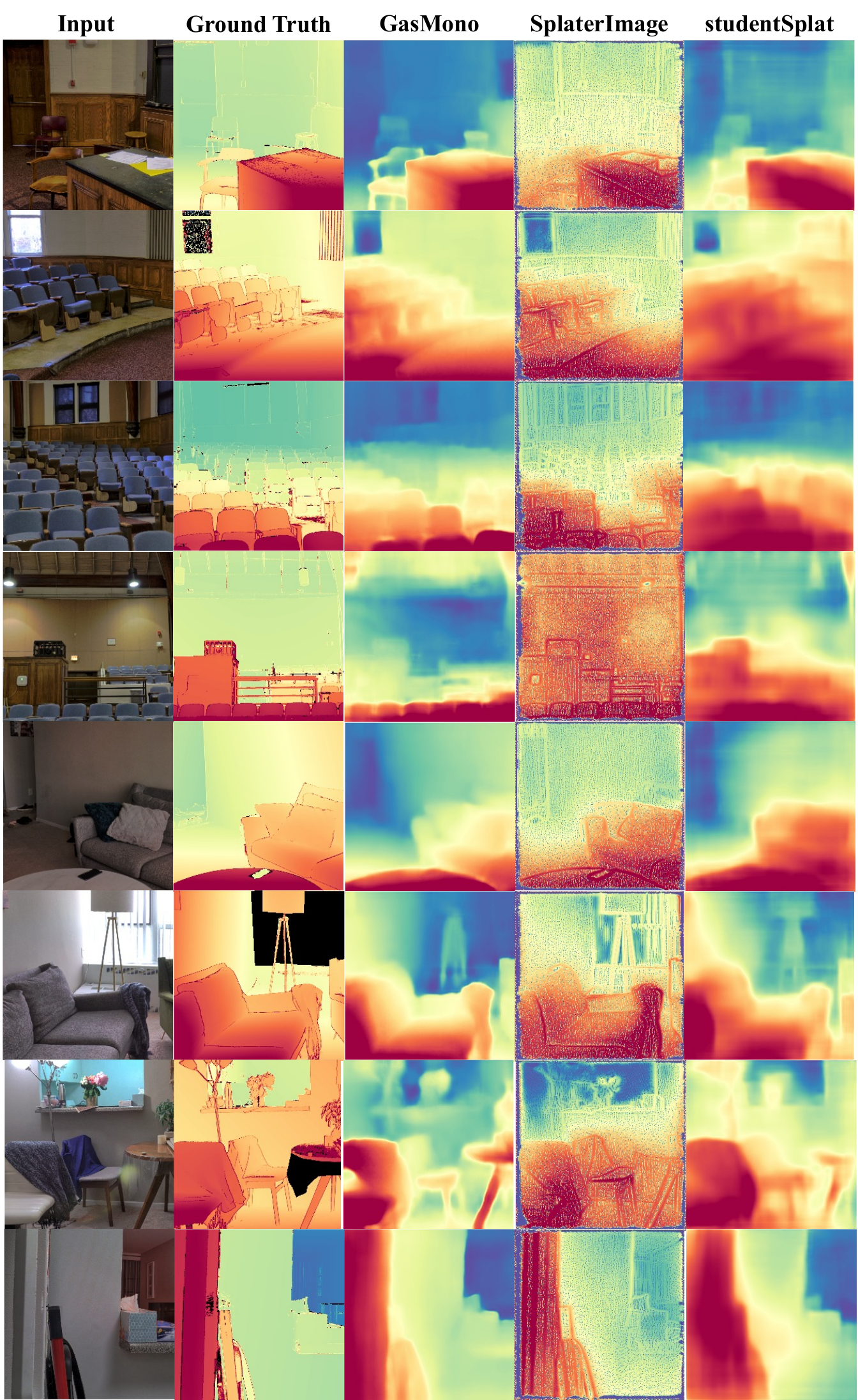}
    \caption{Additional qualitative comparison between representative methods for self-supervised single-view depth estimation performance on the DIODE indoor dataset.}
    \label{fig:add_diode_indoor_qualitative}
\end{figure}

\begin{figure}[t!]
    \centering
    \includegraphics[width=0.9\textwidth]{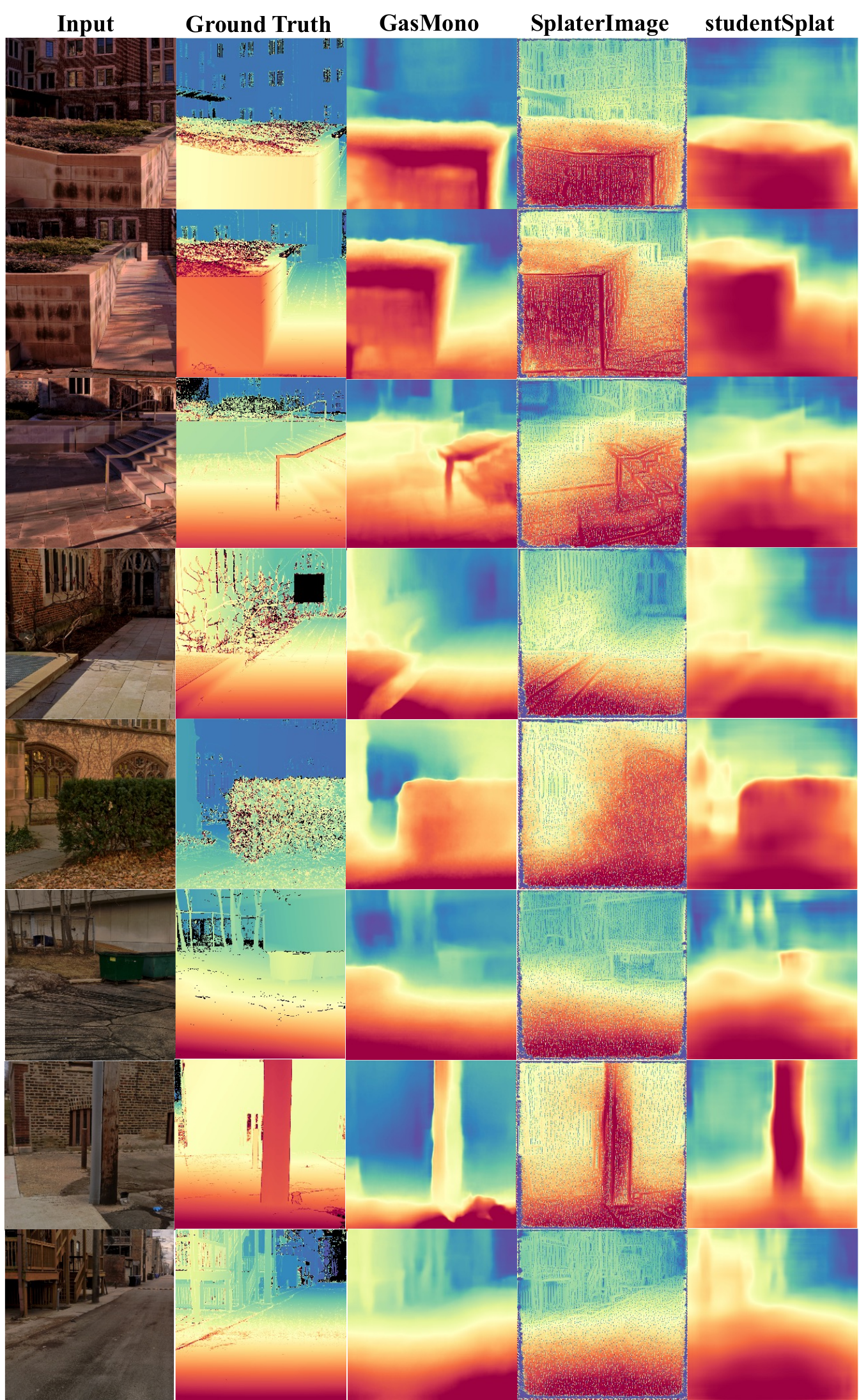}
    \caption{Additional qualitative comparison between representative methods for self-supervised single-view depth estimation performance on the DIODE outdoor dataset.}
    \label{fig:add_diode_outdoor_qualitative}
\end{figure}

\textbf{Qualitative results.} Additional novel-view reconstructions are shown in Figure~\ref{fig:add_re10k_acid_qualitative}. The extrapolating region have lower quality and different content compare to the ground truth. Single-view results can be slightly less sharp. Qualitative results of self-supervised single-view depth estimation are visualized in Figures~\ref{fig:da2k_qualitative}, \ref{fig:add_diode_indoor_qualitative}, and \ref{fig:add_diode_outdoor_qualitative} for the DA2K~\citep{yang2024depth}, DIODE indoor~\citep{vasiljevic2019diode}, and DIODE outdoor~\citep{vasiljevic2019diode} datasets, respectively. Our studentSplat produces less noise compared to SplatterImage~\citep{szymanowicz2024splatter} and is comparable to GasMono~\citep{zhao2023gasmono}. The context confidence weight matrix $\bm W$ is visualized in Figure~\ref{fig:confidence_weight_mat}. The darker regions are less confident, while the brighter regions are more confident. We also show the thresholded $\bm W$ at 0.5 for better visualization. Note that the model is less confident at regions with missing information and object boundaries, where missing context from occlusion tends to happen. This confidence weight guides the extrapolator in our studentSplat model.

\begin{figure}[t!]
    \centering
    \includegraphics[width=0.9\textwidth]{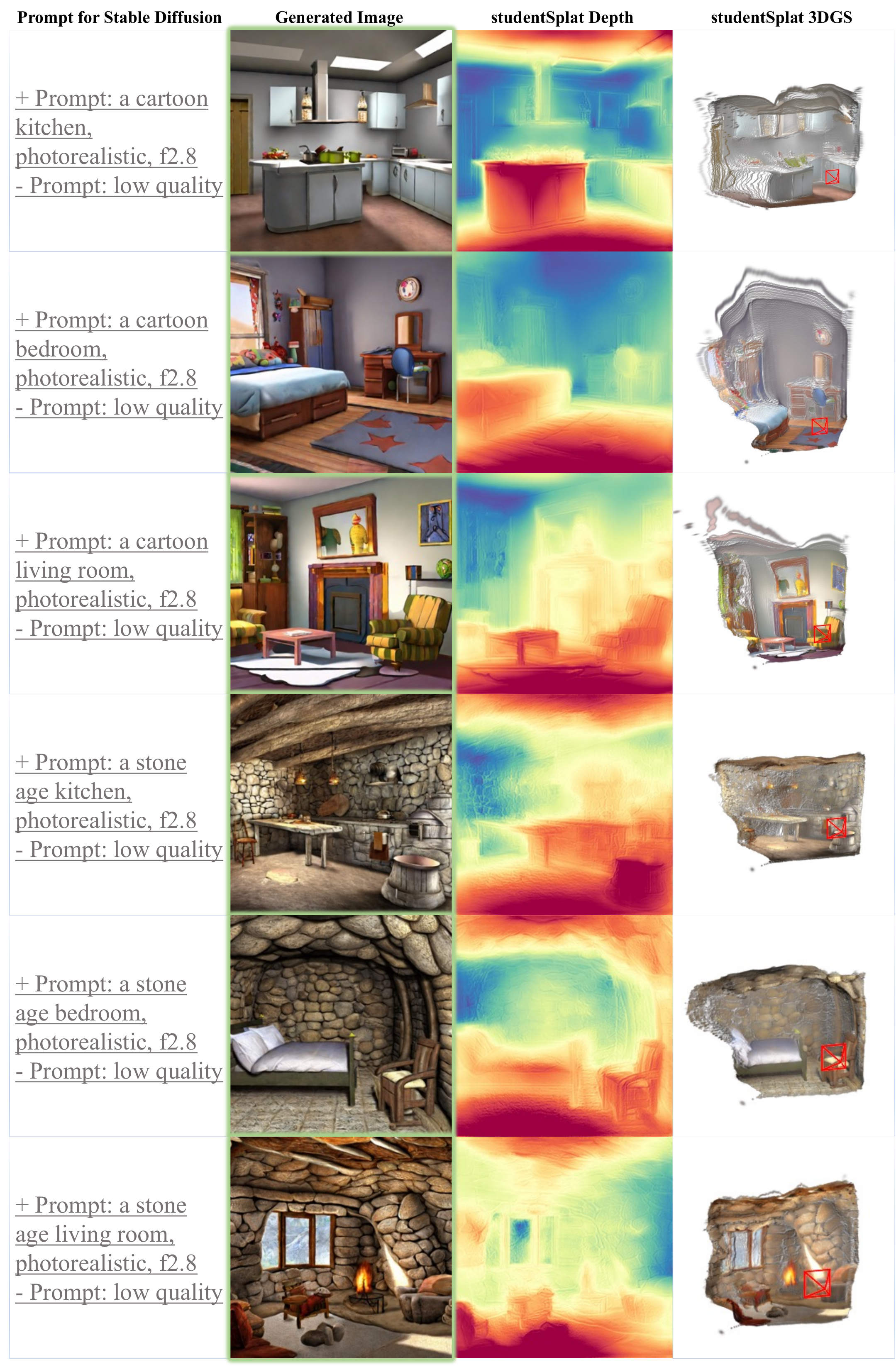}
    \caption{Visualization of the text-to-3D generation result using studentSplat with teacher refine detailed in Section~\ref{sec:teacher_refine} on the Stable Diffusion output. The input is highlighted in green.}
    \label{fig:text_to_3d}
\end{figure}

\textbf{Scene-level text-to-3D generation pipeline.} Generating new 3D views is helpful for 3D design and content creation. Current methods on text-to-3D scene generation require per-scene optimization~\citep{zhang2024text2nerf}, multiple iterations and depth refinement~\citep{ouyang2023text2immersion,fridman2024scenescape,zhang20243d}, or are constrained by predefined objects~\citep{chang2014learning}. By combining studentSplat with Stable Diffusion~\citep{rombach2022high}, we can produce a scene-level text-to-3DGS method that generates diverse 3D scenes without depth guidance. More importantly, we can obtain a text-to-3D scene pipeline without training a 3D generative model. The results are shown in Figure~\ref{fig:text_to_3d}. We apply a fake forward camera shift of 0.2 and use the intrinsics from the training data.

\clearpage
\section{Refining Student Output with Teacher Model} 
\label{sec:teacher_refine}
\subsection{Method}
\begin{figure}[ht!]
    \centering
    \includegraphics[width=\textwidth]{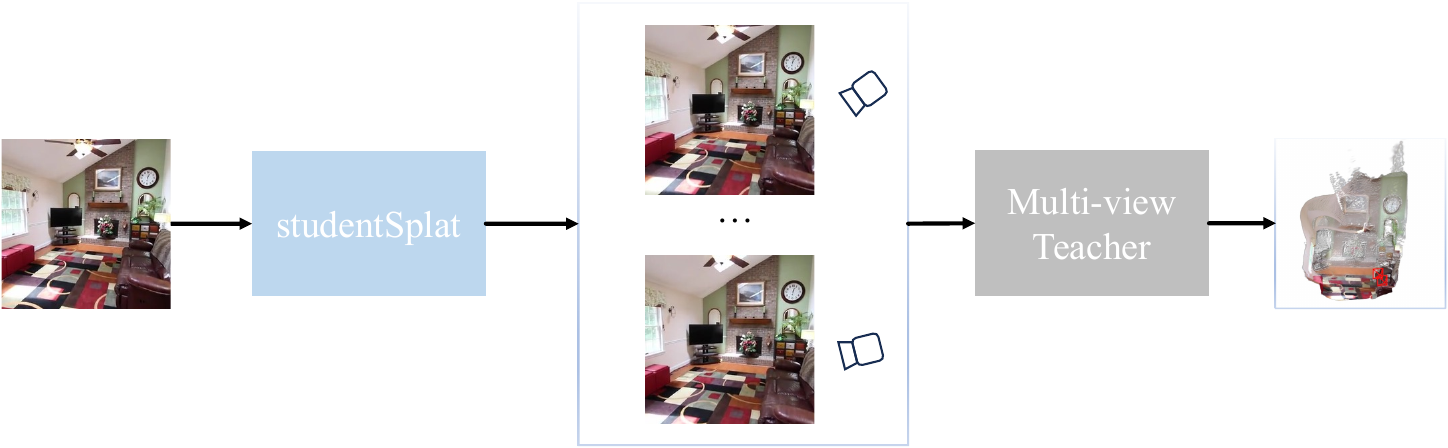}
    \caption{\textbf{The pipeline to refine the student output with the teacher model.} The student model generates additional viewpoints using user-specified virtual camera poses. The teacher model utilizes these generated viewpoints and the corresponding virtual camera poses to refine the camera pose estimates.}
    \label{fig:student_teacher_refine}
\end{figure}

The single-view studentSplat generally produces better quality novel-view reconstructions and extrapolation when the camera view change is small. More importantly, the multi-view teacher model still performs better in 3D reconstruction than the student model. These properties lead us to another design that further improves the 3D reconstruction performance. Specifically, we use studentSplat to generate good quality novel views using one input view and fake camera poses with small shifts. Then, the input view and generated novel views with the fake camera poses are used as the context for the teacher input. The advantage of this pipeline is that we preserve the single-view nature of our studentSplat and only trade off the inference speed for performance improvements. The overall pipeline is shown in Figure~\ref{fig:student_teacher_refine}.
\subsection{Results}
\begin{figure}[ht!]
    \centering
    \includegraphics[width=0.8\textwidth]{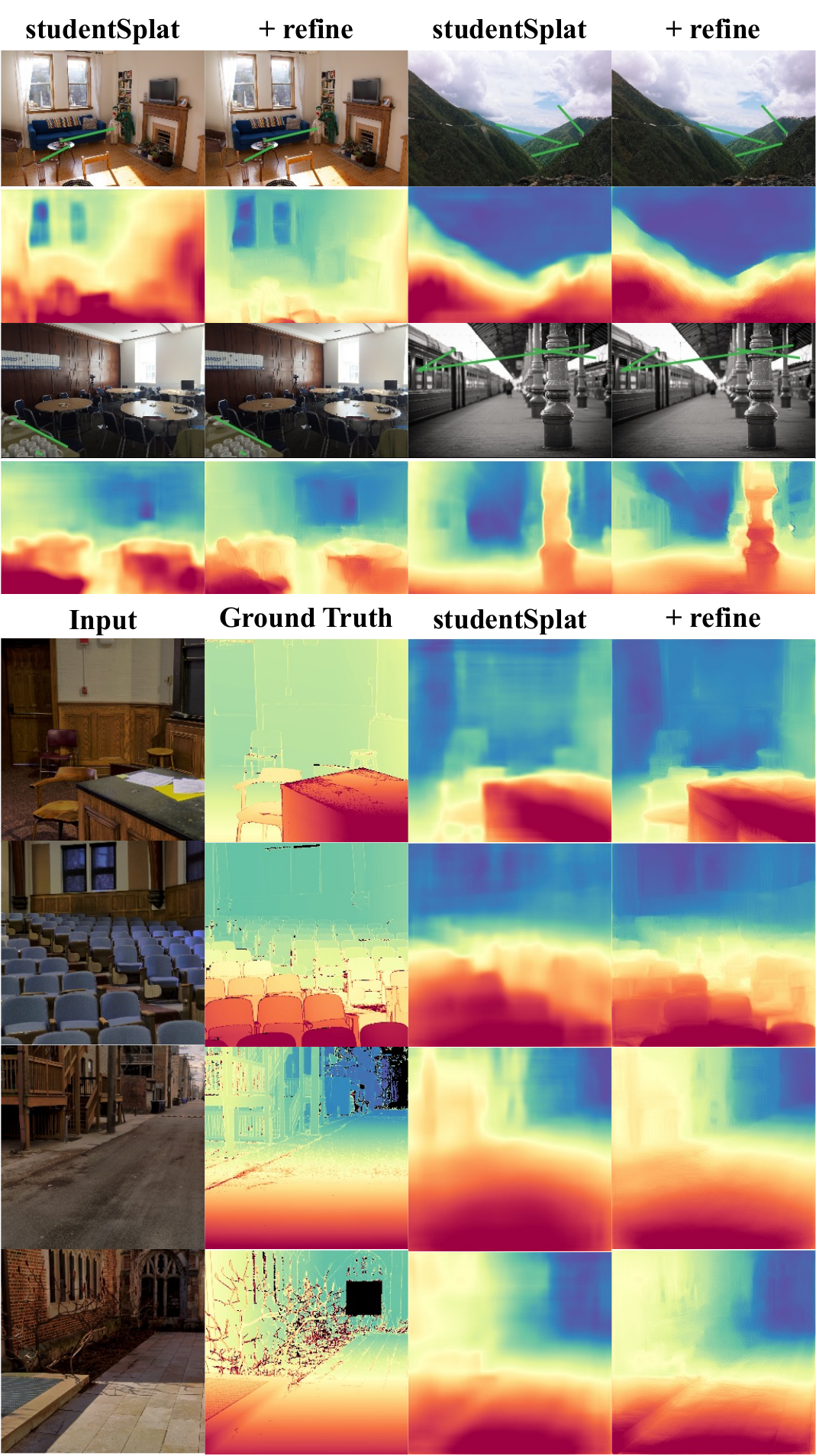}
    \caption{The qualitative comparison between the studentSplat with and without teacher refinement.}
    \label{fig:teacher_refine_qualitative}
\end{figure}

Using the teacher refinement, we can improve the quality of the generated 3D structure. We show the improvements using the single-view depth estimation task. We use a forward ($z$-axis) shift of 0.5 to produce the relative camera poses. All the camera intrinsics, near plane, and far plane are directly taken from the training dataset RE10K~\citep{zhou2018stereo}. Only one image is provided to the pipeline to predict the depth. As shown in Table~\ref{tab:mono_depth_teacher_refine}, the additional use of teacher refinement results in noticeable performance improvements. We can also see from Figure~\ref{fig:teacher_refine_qualitative} that the refined depth maps are much sharper.
\begin{table*}[ht!]
    \begin{center}
\footnotesize
    \setlength{\tabcolsep}{2.0pt} %
    \begin{tabular}{@{}lccc c@{}}
    \toprule
    \multirow{2}{*}[-2pt]{Method} & \multicolumn{2}{c}{DIODE~\scriptsize{\cite{vasiljevic2019diode}}}& & DA-2K~\scriptsize{\cite{yang2024depth}} \\
    \addlinespace[-12pt] \\
    \cmidrule{2-3} \cmidrule{5-5} 
    \addlinespace[-12pt] \\
    & $\delta_1$$\uparrow$ & AbsRel$\downarrow$ & & Acc (\%)$\uparrow$ \\
    \midrule
    % LeReS (Supervised)~\cite{yin2021learning}  & 0.751 & 0.287 && - \\
    % \midrule
    
    GasMono~\scriptsize{\cite{zhao2023gasmono}}  & {0.504} & \textbf{0.348} && {0.700} \\
    \cellcolor{Gray}{studentSplat} & \cellcolor{Gray}\underline{0.604} & \cellcolor{Gray}{0.407} &\cellcolor{Gray}& \cellcolor{Gray}\underline{0.708} \\
    \cellcolor{Gray}{studentSplat w/ teacher refine} & \cellcolor{Gray}\textbf{0.623} & \cellcolor{Gray}\underline{0.397} &\cellcolor{Gray}& \cellcolor{Gray}\textbf{0.716} \\
    \bottomrule
    \end{tabular}
    \end{center}
    \caption{\textbf{Cross-dataset generalization in self-supervised single-view depth estimation.} The studentSplat is trained on the RealEstate10K dataset. "Teacher refine" refers to the additional use of the teacher network to refine the output of the student model.
    }
    \label{tab:mono_depth_teacher_refine}
    \vspace{-0.4cm}
\end{table*}

\end{document}